%% file: main.tex
\documentclass{article}
\usepackage{arxiv}

\usepackage{subcaption}
\usepackage{multirow}
\usepackage{float}
\usepackage{hyperref}       
\usepackage{url}            
\usepackage{booktabs}       
\usepackage{amsfonts}       
\usepackage{nicefrac}       
\usepackage{microtype}      
\usepackage{lipsum}
\usepackage{multirow}
\usepackage{amsmath,amssymb}
\usepackage{subcaption}
\usepackage{graphicx}

\title{Temporal Interpolation of Geostationary Satellite Imagery with Task Specific Optical Flow}

\author{
  Thomas Vandal \\
  NASA Ames Research Center\\
  Bay Area Environmental Research Institute\\
  Moffett Field, CA, United States \\
  \texttt{thomas.vandal@nasa.gov} \\
   \And
  Ramakrishna Nemani\\
  NASA Ames Research Center\\
  Moffett Field, CA, United States \\
  \texttt{rama.nemani@nasa.gov} \\
}

\begin{document}

\maketitle


\begin{abstract}
    Applications of satellite data in areas such as weather tracking and modeling, ecosystem monitoring, wildfire detection, and land-cover change are heavily dependent on the trade-offs to spatial, spectral and temporal resolutions of observations. In weather tracking, high-frequency temporal observations are critical and used to improve forecasts, study severe events, and extract atmospheric motion, among others. However, while the current generation of geostationary satellites have hemispheric coverage at 10-15 minute intervals, higher temporal frequency observations are ideal for studying mesoscale severe weather events. In this work, we apply a task specific optical flow approach to temporal up-sampling using deep convolutional neural networks. We apply this technique to 16-bands of GOES-R/Advanced Baseline Imager mesoscale dataset to temporally enhance full disk hemispheric snapshots of different spatial resolutions from 15 minutes to 1 minute. Experiments show the effectiveness of task specific optical flow and multi-scale blocks for interpolating high-frequency severe weather events relative to bilinear and global optical flow baselines. Lastly, we demonstrate strong performance in capturing variability during a convective precipitation events.
\end{abstract}


\input{introduction}
\input{goesr}
\input{related-work}
\input{methods}
\input{results}
\input{conclusion}

\bibliographystyle{plain}
\bibliography{references}

\newpage

\input{supplement}

\end{document}

%% file: introduction.tex
\section{Introduction}


Every second satellites around the earth are generating valuable data to monitor weather, land-cover, infrastructure, and human activity. Satellite sensors capture reflectance/radiance intensities at designated spectral wavelengths, spatial, and temporal resolutions. Properties of the sensors, including wavelengths and resolutions, are optimized for particular applications. Most commonly, satellites are built to capture the visible wavelengths, which are essentially RGB images.  Scientific specific sensors capture a larger range of wavelengths, such as micro, infrared, and thermal waves, providing information to many applications such as storm tracking and wildfire detection. However, sensing a greater number of wavelengths is technologically more complex and applies further constraints of temporal and spatial resolution. Similarly, a higher temporal frequency requires high altitude orbital dynamics which then affects the spatial resolution due to it's distance from earth.

NASA and other agencies have developed satellites designed for a variety of applications in both polar and geostationary orbits. Polar orbiting satellites cross south and north poles each revolution around the earth. These satellites have relatively low altitude orbits which allow for high spatial resolution but with an optimal revisit interval of 1-day. NASA's Moderate Resolution Imaging Spectroradiometer (MODIS)~\cite{pagano1993moderate} and Landsat-8~\cite{roy2014landsat} satellites follow a polar orbit with 1- and 8-day revisit times, respectively. Data provided by MODIS and Landsat are widely used for quantifying effects of climate change, land-cover usage, and air pollution, among others, but are not well suited to monitoring high-frequency events. On the other hand, geostationary satellites are well suited for sub-daily events such as tracking weather events and understanding diurnal cycles. The geostationary orbit keeps satellites in a consistent point $35,786$km above Earth's equator. While the high altitude reduces spatial resolution, the current generation of geostationary satellites is able to provide minute-by-minute data, enabling immense opportunity for understanding atmospheric, land-cover, and oceanic dynamics. 

Within a few years, a constellation of geostationary satellites by multiple international institutions will provide global coverage of earth's state. Latest generation of geostationary satellites includes NOAA/NASA's GOES-16/17~\cite{goesdocs}, Japan's Himwari-8/9~\cite{bessho2016introduction}, China's Fengyun-4~\cite{yang2017introducing}, and Korea's GEO-KOMPSAT-2A with future plans in development. Full-disk coverage from such satellites have revisit times of 10-15 minutes allowing applications to real-time detection and observation of wildfires~\cite{xu2017real}, hurricane tracking, air flood, precipitation estimation, flood risk, and others~\cite{schmit2018applications}.  Further, given improved spectral and spatial resolution in current generation sensors, geostationary satellites open opportunities to incorporate and learn from less frequent observations from polar orbiters.

While 10-15 minute revisit times is temporally sufficient for many applications, higher frequency snapshots can aid a variety of tasks. For instance, understanding rapidly evolving convective events is a high priority for improving atmospheric models, which are notoriously poor at simulating heavy precipitation and as highlighted in NASA's Earth Science Decadal Survey \cite{board2019thriving}. However, data for analyzing such events is often not available at the desired frequency. Similarly, comparing multiple satellite observations is dependent on their corresponding timestamps. This leads to an interpolation task between observations in a multi-spectral spatio-temporal sequence, similar to that of video interpolation.

Optical flow is a problem of tracking apparent motion by estimating partial derivatives between images. Optical flow is the basis for top performing video interpolation methods~\cite{liu2019deep,bao2019depth,jiang2018super}. In this work, we adapt Superslomo (SSM)~\cite{jiang2018super}, a video interpolation method, to the problem of temporal interpolation between geostationary images. We compare properties of global, task specific, and multi-scale block SSM models with the traditional linear interpolation. Our experiments show that task specific SSM is capable of interpolating high-frequency severe atmospheric events. Further, visual analysis suggests the learned optical flows resemble atmospheric with dynamic visibility maps. 


The remainder of this paper is outlined as follows. Section 2 discusses related work including resolution enhancement in the earth sciences and the current state of video intermediate frame interpolation. Section 3 introduces the GOES-R dataset and section 4 details the methodology. Experiments on a large scale dataset and a severe storm case study are presented in section 5. Lastly, section 6 concludes with challenges and further work.

\begin{figure}
     \centering
     \begin{subfigure}{0.49\linewidth}
         \centering
         \includegraphics[width=\linewidth]{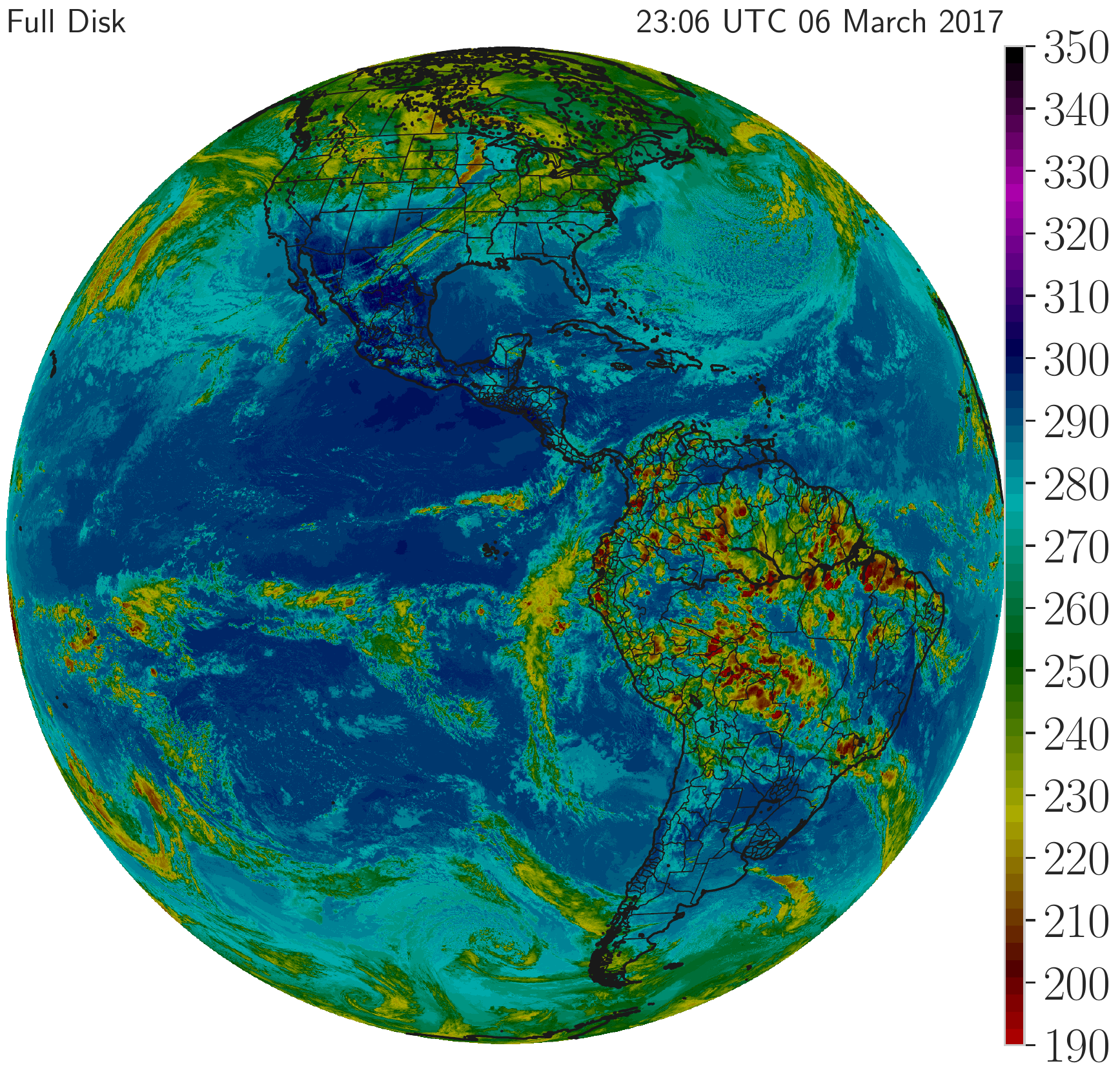}
         \caption{}
         \label{fig:ir_fd_map}
     \end{subfigure}
     \begin{subfigure}{0.49\linewidth}
         \centering
         \includegraphics[width=\linewidth]{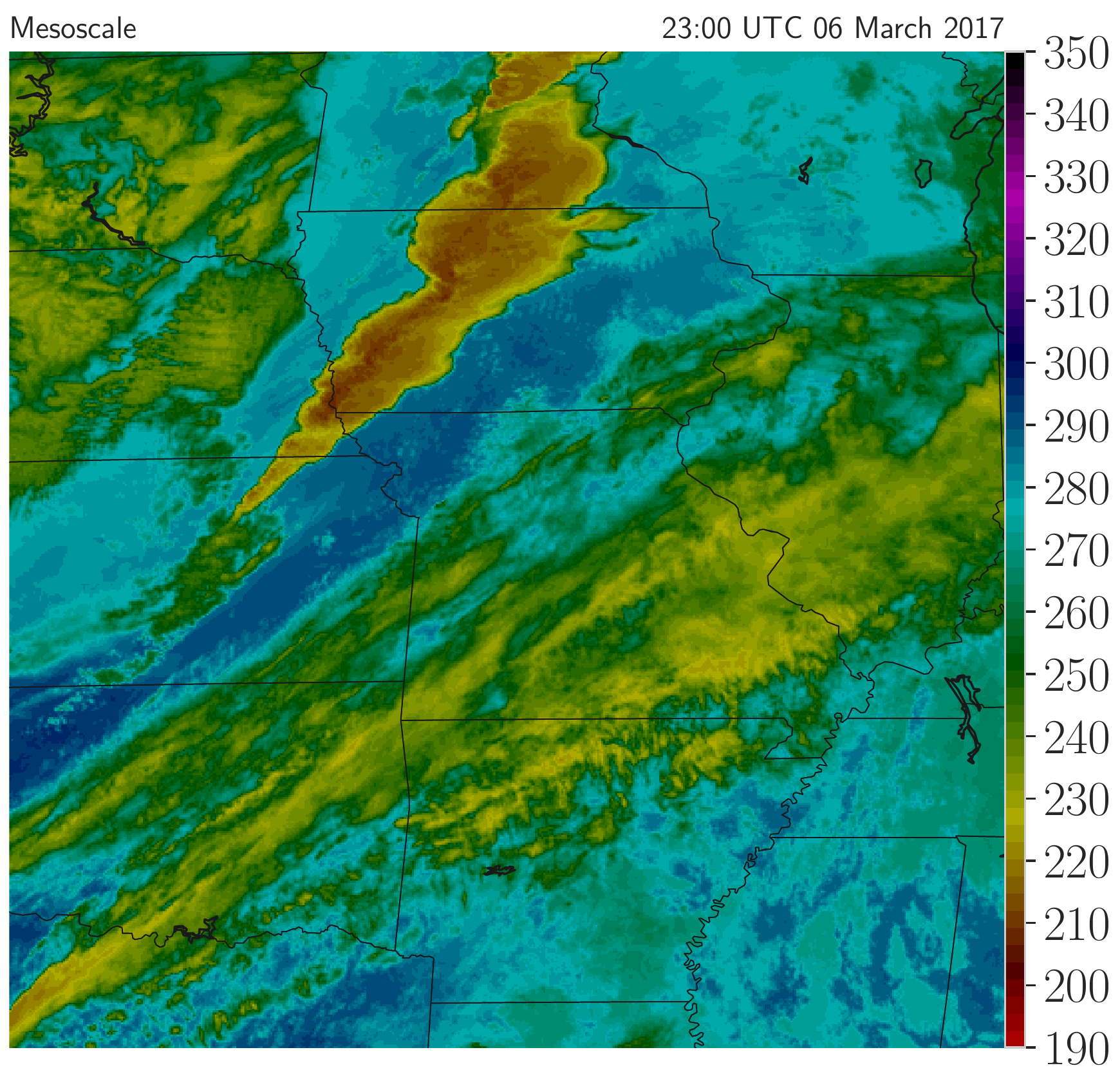}
         \caption{}
         \label{fig:ir_meso_map}
     \end{subfigure}
    \caption{Cloud-top temperature from "clean" IR long-wave window Band 13 Full-disk (\ref{fig:ir_fd_map}) and Mesoscale (\ref{fig:ir_meso_map}) coverage.}
    \label{fig:ir-maps}
\end{figure}

%% file: goesr.tex
\section{GOES-R Satellite Dataset} \label{sec:goesr}

Geostationary satellites are synchronized in orbit with earth's spin to hover over a single location. Given this location, the sensor, measuring radiation as often as possible, can frequently capture data over a continuous and large region. This feature makes geostationary satellites ideal for capturing environmental dynamics. The GOES-R series satellites, namely GOES-16/17 (East and West side of the Americas), operated by NASA and NOAA provides scientists with unprecedented temporal frequency enabling real-time environmental monitoring using the Advanced Baseline Imager (ABI)~\cite{schmit2017closer}. GOES-16/17 senses 16-bands of data which are listed in Table \ref{tab:goesr} with central central wavelength, spatial resolution, and band name. Three data products are derived from each GOES-16/17; 1. Full-disk covering the western hemisphere every 15-minutes (figure \ref{fig:ir_fd_map}), 2. Continental US every 5-minutes, and 3. Mesoscale user directed 1000km by 1000km sub-region every at an optimal 30 seconds (figure \ref{fig:ir_meso_map}). ABI's 16 spectral bands includes two visible (1-2), four near-infrared (3-6), and ten infrared (7-16) bands enabling a suite of applications. 

These geostationary satellites are particularly useful in tracking weather, monitoring high-intensity events, estimating rainfall rates, fire detection, and many others at near real-time. Mesoscale mode gives forecasters the ability to ``point'' the satellite at a user specific sub-region for near constant monitoring of severe events. For example, GOES-16 provided emergency response units tools for decision making during the 2018 California wildfires. However, this high frequency data also provides valuable information of environmental dynamics and retrospective analysis, such as studying convective events~\cite{fiolleau2013algorithm}. Furthermore, mesoscale data can be used to inform techniques to produce higher temporal resolution CONUS and full-disk coverage. In this work, we develop a model to improve the temporal resolutions of CONUS and full-disk by learning an optical flow model to interpolate between consecutive frames. With this, we are able to generate 1-minute full-disk artificially enhanced data. 

\begin{table}
    \centering
    \begin{tabular}{c|c|c|l}
        \toprule
         \multirow{3}{0.08\linewidth}{\centering Band} &  \multirow{3}{0.19\linewidth}{\centering Central Wavelength ($\mu$m)} & \multirow{3}{0.18\linewidth}{\centering Spatial Resolution (km)}& \multirow{3}{0.2\linewidth}{\centering Name} \\
         & & & \\
         & & & \\
         \midrule
         1 & 0.47 & 1 & Blue \\
         2 & 0.64 & 0.5 & Red \\
         3 & 0.86 & 1 & Veggie \\
         4 & 1.37 & 1 & Cirrus \\
         5 & 1.6 & 1 & Snow/Ice \\
         6 & 2.24 & 2 & Cloud Particle Size \\
         7 & 3.9 & 2 & Shortwave Window \\
         8 & 6.2 & 2 & Upper-level Water Vapor \\
         9 & 6.9 & 2 & Mid-level Water Vapor \\
         10 & 7.3 & 2 & Low-Level Water Vapor \\
         11 & 8.4 & 2 & Cloud-Top Phase \\
         12 & 9.6 & 2 & Ozone \\
         13 & 10.3 & 2 & "Clean" IR Longwave \\
         14 & 11.2 & 2 & IR Longwave \\
         15 & 12.3 & 2 & "Dirty" IR Longwave \\
         16 & 13.3 & 2 & CO$_2$ Longwave IR \\
         \bottomrule
    \end{tabular}
    \caption{GOES-R Series Bands}
    \label{tab:goesr}
\end{table}

%% file: related-work.tex
\section{Related Work}

In this section we begin by reviewing previous work in the areas of data fusion and resolution enhancement as applied generally to remote sensing satellite imagery as well as some recent successes of deep learning in the area. Secondly, we provide a brief review of video intermediate frame interpolation techniques.

\subsection{Resolution Enhancement of Satellite Data}

Earth science datasets are complex and often require extensive preprocessing and domain knowledge to effectively render itself useful for large-scale applications or monitoring.  Such datasets may contain frequent missing values due to sensor limitations, low quality pixel intensities, incomplete global coverage, and contaminated with atmospheric processes related to cloud and aerosols. Further, spatial and temporal resolution enhancement is often applied to improve analysis precision. Techniques to handle these challenges have been developed and are widely applied across the remote sensing community. 

Many statistical and machine learning methodologies for improving spatial resolution have been explored and is an active area of research. Data fusion is one area where two or more datasets are \textit{fused} to generate an enhanced product, often with both higher spatial and temporal resolutions~\cite{hall1997introduction}. The Spatial and Temporal Adaptive Reflectance Fusion (STARFM) algorithm, for example, uses Landsat and MODIS to produce a daily 30-meter reflectance product by using a spectral wise weighting model~\cite{zhu2010enhanced}. Similarly, nearest neighbor analog multiscale patch-decomposition data driven models are used as state-of-the-art interpolation techniques for developing global sea surface temperature (SST) datasets~\cite{fablet2017data}. In recent years, super-resolution techniques have presented state-of-the-art results for spatial enhancement of satellite images \cite{yang2015remote,li2017super,lanaras2018super}. 

Approaches for temporal resolution enhancement of individual satellite observations have not been as well studied. Liebmann et al. presented the first linearly interpolated datasets filling in missing and erroneous longwave radiation many days apart to improve global coverage~\cite{liebmann1996description}. Similarly, \cite{kandasamy2013comparison} presented a comparison of multiple methods for interpolating between MODIS observations to generate a synthetic leaf area index dataset. A number of statistical techniques including long-term climatology measures and time-series decomposition were applied to smooth observation and fill gaps. \cite{doelling2013geostationary} presented an approach using linear interpolation on sub-daily geostationary imagery to match timestamps between multiple satellites. However, given more frequent observations by the recent generation of geostationary observations, more complex methods beyond linear interpolation may be more applicable and accurate in the temporal domain.

Our work proposes to apply deep learning methodologies to optimize the interpolation problem. In recent years, a number of applications in processing and learning from satellite data have shown state-of-the-art results using deep learning. For example, \cite{benedetti2018m} showed that recurrent and convolutional neural networks effectively assimilate multiple satellite images. \cite{lanaras2018super} presented a global deep learning super-resolution approach for Sentinal-2 with a 50\% improvement beyond traditional techniques. In terms of classification, DeepSat showed that normalized deep belief networks tuned where able to outperform traditional techniques for image classifications~\cite{basu2015deepsat}. Convolutional neural networks have been shown to effectively classify land use in remotely sensed images, from urban areas~\cite{castelluccio2015land} to crop types~\cite{kussul2017deep}. 

While many studies have explored resolution enhancement spatially, and temporally, the authors are not aware of any prior work exploring temporal interpolation at the minute-to-minute scale. Prior approaches on longer time scales have applied linear interpolation and nearest neighbor techniques. We will explore the applicability of a more complex optical flow approach to temporal interpolation at very high resolutions and use linear interpolation as our baseline, as applied in prior work. 

\subsection{Video Intermediate Frame Interpolation}

Video interpolation techniques have shown high skill at generating slow motion footage by generating intermediate frames in spatially and temporally coherent sequences~\cite{liu2019deep,bao2019depth,liu2017video,jiang2018super}. These approaches are designed to learn the dynamics by inferring displacement of spatial structure between consecutive images. Optical flow is widely used for this task which estimates pixel displacement by comparing two images and interpolating appropriately. For RGB imagery, this task is equivalent to estimating movement of objects.  In recent years, deep learning architectures have shown promising results for both optical flow and video interpolation. Flownet presented an encoder-decoder architecture for optical flow with correlation operations and skip-connections for supervised learning~\cite{dosovitskiy2015flownet}. Following studies have extended this work with more complex architectures such as stacking networks for large and small displacements \cite{ilg2017flownet} and unsupervised learning \cite{meister2018unflow}. Deep Voxel Flow \cite{liu2017video}, Superslomo \cite{jiang2018super}, cyclic frame generation \cite{liu2019deep}, and others have shown deep learning optical flow techniques to be well suited for video interpolation.


However, many video interpolation techniques focus on \textit{single frame} interpolation, meaning that a single frame is estimated between two consecutive frames~\cite{niklaus2017video,liu2017video,liu2019deep}. However, when interpolating satellite imagery, time-dependent and multi-frame estimation is preferred for more flexibility. Jiang et al. presented Superslomo (SSM) which combines both optical flow and occlusion models for time-dependent estimation between consecutive frames~\cite{jiang2018super}. The time-dependent nature of this approach produces spatially and temporally coherent predictions of any time between 0 and 1. In their experiments, ~\cite{jiang2018super} shows that 240-fps video clips can be estimated from 30-fps inputs. Further details of this work will be presented in Section 4 where we apply their architecture with an extension to multi-scale optical flows. 


High-frequency satellite imagery can take advantage of these techniques to extract dynamics of different physical processes. We study how SSM can be effectively applied to this problem by experimenting with global and task specific models.


%% file: methods.tex
\section{Methodology}

    Temporal up-sampling of geostationary satellite data is a near identical problem as intermediate video frame interpolation with domain specific characteristics. In video interpolation, the goal is to estimate an intermediate frame given two or more consecutive RGB images. A single set of optical flows are sufficient for interpolating between RGB images as objects captured in the visible spectrum are reasonably consistent across frames. However, as discussed above, satellite imagery often consists of 10's or even 100's of spectral channels with varying spatial resolutions. Further, each channel captures different physical properties with heterogeneous motion including severe events such as convection leading to heavy precipitation and tornadoes. The goals of the proposed methodologies include interpolating to a user defined point in time, capturing varying spatial dynamics, and computational efficiency at scale. In this section, we review the SSM framework for temporal up-sampling with optical flow, as presented by~\cite{jiang2018super}. Next, we discuss task specific SSM models and the chosen network architecture with multi-scale blocks.

    \begin{figure} 
        \begin{subfigure}{0.49\linewidth}
            \centering
            \includegraphics[width=\linewidth]{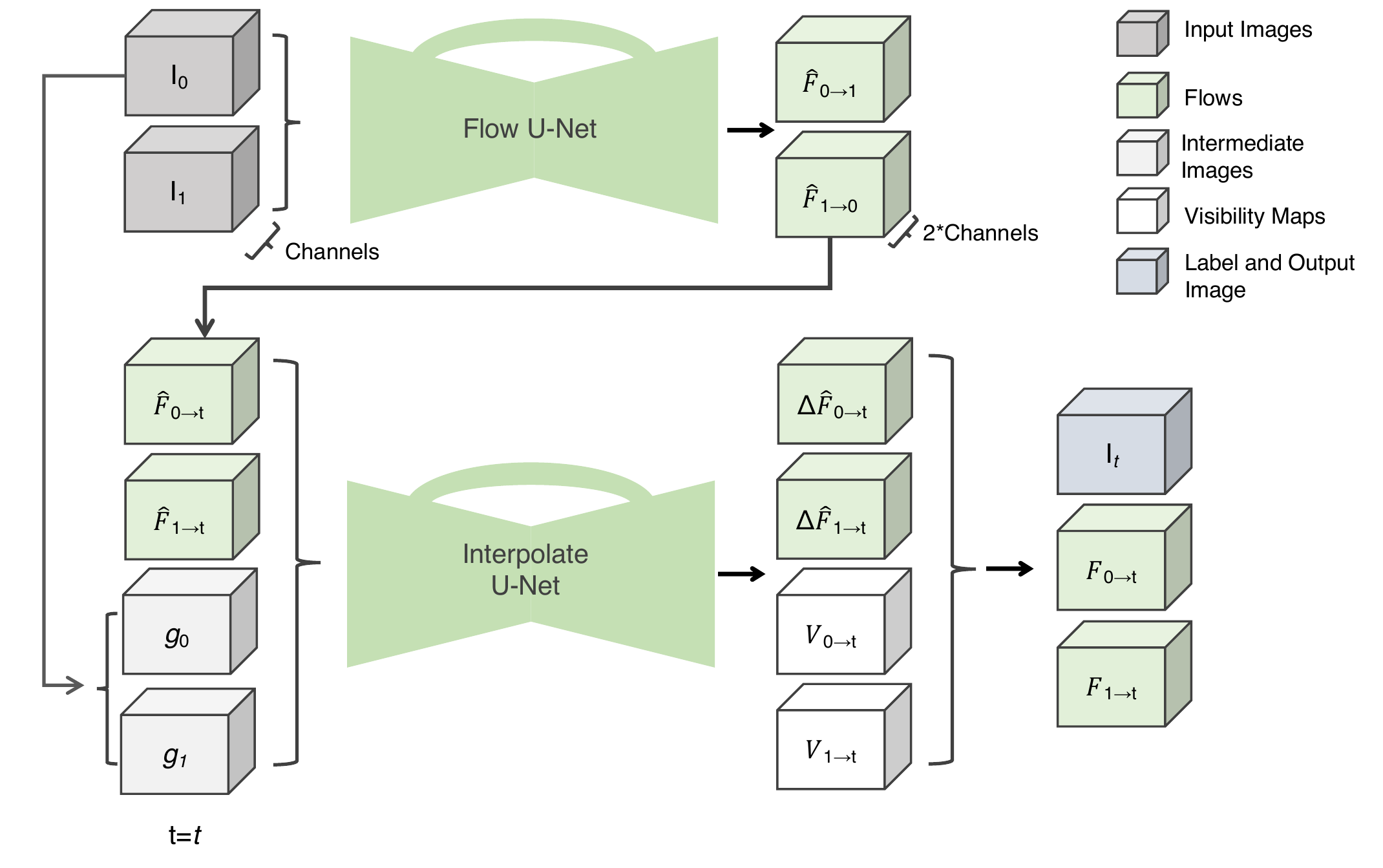}
            \caption{SuperSlomo Architecture with Flow and Interpolation Networks.}
            \label{fig:slomo}
        \end{subfigure}
        \begin{subfigure}{0.49\linewidth}
            \centering
            \includegraphics[width=\linewidth]{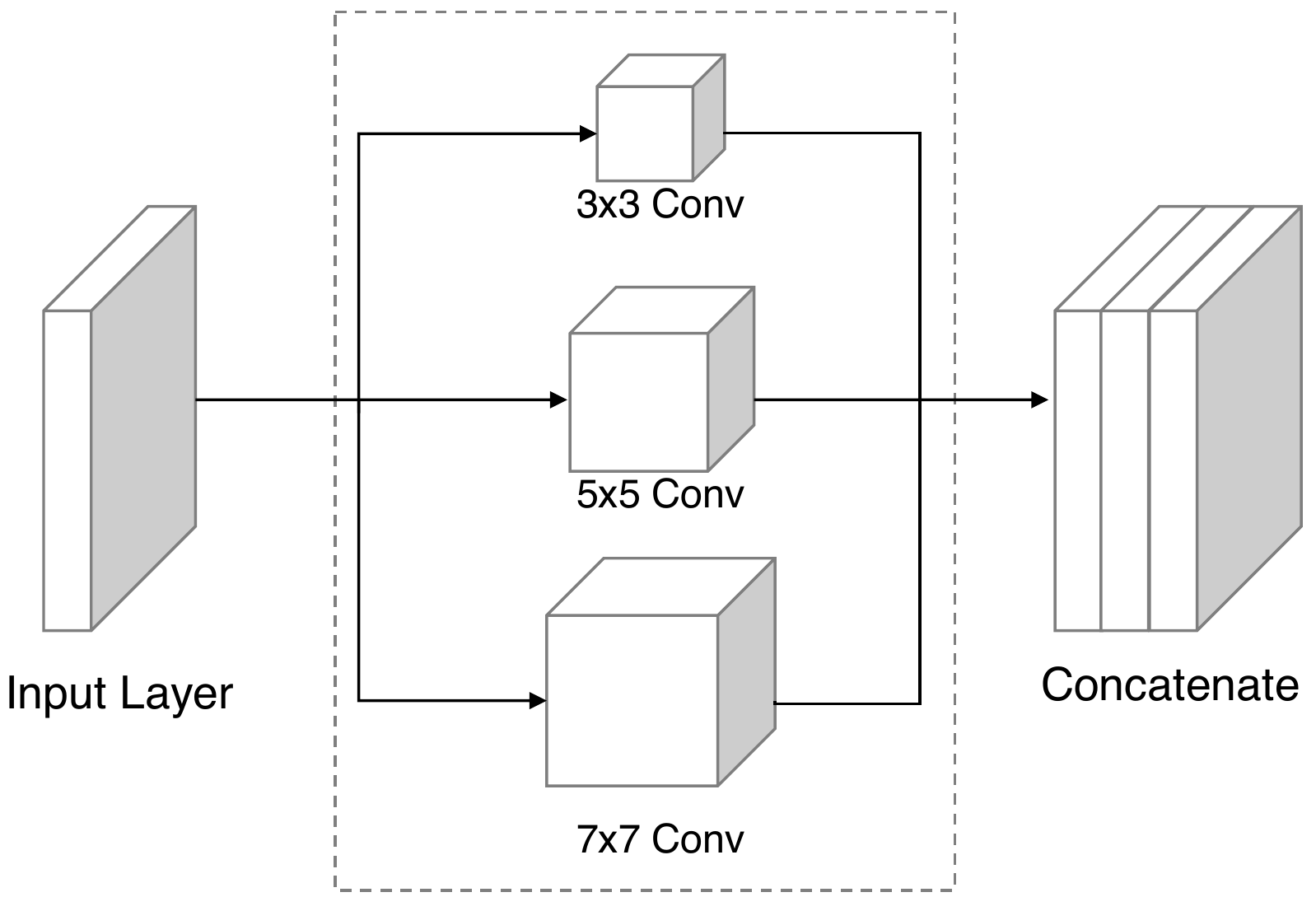}
            \caption{Multi-scale block}
            \label{fig:multiscale}
        \end{subfigure}
    \end{figure}
\subsection{Intermediate Frame Interpolation}
    
    SSM intermediate frame interpolation considers the case of frame estimation at a user defined point in continuous time \cite{jiang2018super}. To ensure smooth transitions and structural similarity between frames, SSM is designed to predict optical flows between two input images as a function of time. The approach, which can be seen in Figure~\ref{fig:slomo}, consists of two deep neural networks. The first estimates forward and backward flows between two input images.  The second network, depending on time, updates the forward and backward flows and generates visibility maps to handle occlusion. These features of SSM are well suited to geostationary data by enabling arbitrary temporal up-sampling and synchronization of multiple datasets. 
    
    Following the notation from~\cite{jiang2018super}, let $I_0, I_1, I_t \in \mathcal{R}^{H \times W}$ where $t \in (0,1)$, $H$ as image height and $W$ as image width, and $C$ a number of spectral bands. In our case, $C=1$. The goal is then to construct an intermediate frame $I_{t}$ with a linear combination of warped $I_{0}$ and $I_{1}$ as defined by:
    
    \begin{equation}
        \hat{I}_{t} = \alpha \cdot g(I_{0}, F_{0 \rightarrow t}) + (1-\alpha) \cdot g(I_{1}, F_{1 \rightarrow t})
        \label{eq:general-interpolation}
    \end{equation}
    
    \noindent where $F_{0 \rightarrow t}$ and $F_{t \rightarrow 1}$ are the optical flows from $I_{0}$ to $I_{t}$ and $I_{t}$ to $I_{1}$, respectively. 
    $g$ is defined as the \textit{backward warping} function, implemented with bilinear interpolation, and $\alpha$ represents a scalar weight coefficient to enforce temporal consistency and allow for occlusion reasoning.  
    In the case of high temporal resolution satellite imagery, the interpolation is virtually estimating the state of atmospheric variables (clouds, water vapor, etc.) over a static land surface. 
    If a given pixel in $I_0$ captures land surface but the same pixel in $I_1$ sees a cloud, the occlusion principle is used to estimate at what time $t$ the cloud covers the pixel. 
    Further, atmospheric dynamics cause physical characteristics to change over time. 
    One example is convection such that warm/cold air vertically and rapidly mixes in the atmosphere causing severe weather events. 
    In the context of interpolating, dynamics between $I_0$ and $I_1$ cause cloud temperature to rapidly decrease, leading to a drastic change brightness intensity and breaking assumptions of optical flow. 
    However, visibility maps, $V_{0 \rightarrow t},V_{1 \rightarrow t} \in (0,1)^{H \times W}$, weight brightness importance to account for both occlusion and intensity changes. Equation~\ref{eq:general-interpolation} is then be redefined as:
    
    \begin{equation}
        \hat{I}_{t} = \dfrac{1}{Z} \cdot \big( (1-t) \cdot V_{0 \rightarrow t} \cdot g(I_{0}, F_{0 \rightarrow t}) + t  \cdot V_{1 \rightarrow t} \cdot g(I_{1}, F_{1 \rightarrow t}) \big)
        \label{eq:Ithat} 
    \end{equation}

    \noindent where $Z = (1-t) \cdot V_{0 \rightarrow t} + t \cdot V_{1 \rightarrow t}$ is a normalization factor. 
    Forward and backward optical flows, $(F_{0 \rightarrow t}, F_{0 \rightarrow t})$, at time $t$ are estimated by a sequence of two flow networks, $G_{\text{flow}}$ and $G_{\text{Interp}}$, as presented in Figure~\ref{fig:slomo}. 
    The first network, $G_{\text{flow}}(I_0,I_1)$, infers backward and forward optical flows, $(F_{0 \rightarrow 1},F_{1 \rightarrow 0})$, between two input images. 
    After generating approximate intermediate flows, $(\hat{F}_{t \rightarrow 0}, \hat{F}_{t \rightarrow 0})$, intermediate images are generated. 
    The interpolation network, $G_{\text{Interp}}$, predicts visibility maps $(V_{0 \rightarrow t},V_{1 \rightarrow t})$ and final flows $(F_{t \rightarrow 0}, F_{t \rightarrow 0})$ as a function of a concatenation of input images, intermediate flows, and intermediate warped images. 

\subsection{Task Specific Interpolation}

    As discussed in Section~\ref{sec:goesr}, GOES-16 consists of 16 channels with resolutions between 500m and 2km. Flows between images with different spatial resolutions will have flows of varying intensity. Clouds in 500m images will cover 4x more pixels than a corresponding 2km image. We explore the use of task specific networks by learning separate SuperSlomo models for each spectral channel and compare with a single global model. While requirements for GPU computation multiplies, we will show that improved performance of task specific models improves results substantially.

\subsection{Network Architecture}

    Deep neural networks with encoding and decoding are well suited to model both local and global spatial structure. Architectures of this type include Flownet~\cite{dosovitskiy2015flownet} and U-Net~\cite{jiang2018super} which have been shown to perform well in the task of optical flow. We follow this approach using a U-Net architecture for each of the flow and interpolation networks. 
    The U-Net architecture applied has 4 down-sampling layers followed by 4 up-sampling layers with skip connections between each corresponding layer. A convolution layer maps the input to 64 channels with a kernel size of 7. The following downsampling layers are of size 128, 256, 512, and 512 with kernel sizes 5, 5, 3 and 3. Each downsampling layer performs. average pooling and two convolutions with rectified linear unit (ReLu) activations. Upsampling layers of size 256, 128, 64, and 32 all with kernel sizes of 3 is then applied. Each layer performs bilinear interpolation followed by two convolutions with rectified linear unit (ReLu) activations. Lastly, 32 channels in the last hidden layer are mapped to the number of output channels using a convolution operation of kernel size 3. Flow and interpolation networks use the same architecture with different input and output dimensions as discussed above.
    
    Tracking both small and large displacements continues to be a challenge, even with encoder-decoder network architectures. Other approaches have shown that using a stack of networks performing small and large displacement perform well~\cite{ilg2017flownet}. In this work, we explore the applicability of multi-scale hidden layers to track local and global features. We follow a similar approach applied in ~\cite{ding2019lightweight} where hidden layers are defined to have multiple convolution operations with different sized kernels followed by a concatenation layer, as shown in Figure~\ref{fig:multiscale}. In our networks, kernels of size 3, 5, and 7 conserve high-frequency spatial details while abstracting global motion for improved optical flows and visibility maps. 
    
\subsection{Training Loss}

        As all variables in the architecture are differentiable, the model can be learned in an end-to-end manner. Given two inputs frames $I_0$ and $I_1$ with $N$ intermediate frames $\{I_{t_i}\}_{i=1}^N$ and corresponding predictions $\{\hat{I}_{t_i}\}_{i=1}^N$ a loss function can be defined as a weighted combination of reconstruction, warping, and smoothness losses such that:
        
        \begin{equation}
            l = \lambda_r l_r + \lambda_w l_w + \lambda_s l_s.
            \label{eq:totalloss}
        \end{equation}
        
        \noindent We note that~\cite{jiang2018super} includes a fourth term for perception of image classes which are not available for this satellite dataset. Similarly, we employ $L_1$ loss functions for each loss terms unless noted otherwise. 
        
        The \textit{reconstruction loss} is defined as the distance between observed and predicted intermediate frames:
        \begin{equation}
            l_r = \dfrac{1}{N} \sum_{i=1}^N ||\hat{I}_{t_i} - I_{t_i}||.
        \end{equation}
        
        A \textit{warping loss} is used to optimize estimated optical flows between input and intermediate frames for a channel $c$: 
        \begin{equation}
            \begin{split}
                l_w &= ||I_{0} - g(I_{1}, F_{0 \rightarrow 1})|| + ||I_{1} - g(I_{0}, F_{1 \rightarrow 0})|| + \\
                & \dfrac{1}{N} \sum_{i=1}^N ||I_{t_i} - g(I_{0}, F_{0 \rightarrow t_i}) || +
                 \dfrac{1}{N} \sum_{i=1}^N ||I_{t_i} - g(I_{1}, F_{1 \rightarrow t_i}) ||.
            \end{split}
        \end{equation}

        A \textit{smoothness loss} is applied to forward and backward flows from $I_0$ to $I_1$ to satisfy the smoothness assumption of optical flows in the first network such that:
        \begin{equation}
            l_s = ||\triangledown F_{0 \rightarrow 1}||_1 + ||\triangledown F_{1 \rightarrow 0}||_1
        \end{equation}
        
        In practice, this training setup requires optimization over multiple hyper-parameters including $\lambda_r$, $\lambda_s$, $\lambda_w$, and a learning rate. 

%% file: results.tex
\begin{table*}
    \scriptsize
    \begin{tabular}{l|rrrr|rrrr|rrrr}
    \toprule
    {} & \multicolumn{4}{c}{ PSNR $\uparrow$} & \multicolumn{4}{c}{RMSE $\downarrow$} & \multicolumn{4}{c}{SSIM $\uparrow$} \\
    Model & Linear &  SSM-G &  SSM-T & SSM-TMS & Linear & SSM-G & SSM-T & SSM-TMS & Linear & SSM-G & SSM-T & SSM-TMS \\
    Band &        &        &         &          &        &       &         &          &        &       &         &          \\
    \midrule
    1    & 37.795 & 36.828 & \textbf{38.282} &  37.818 &  0.178 & 0.198 & \textbf{0.160} &   0.185 &  0.719 & 0.682 & \textbf{0.734} &   0.722 \\
    2    & 37.408 & 37.006 & \textbf{37.748} &  37.583 &  0.185 & 0.186 & \textbf{0.169} &   0.177 &  0.637 & 0.616 & \textbf{0.649} &   0.644 \\
    3    & 41.808 & 40.544 & \textbf{41.350} &  41.135 &  0.100 & 0.112 & \textbf{0.099} &   0.108 &  0.760 & 0.712 & 0.731 &   \textbf{0.737} \\
    4    & 60.519 & 60.838 & \textbf{62.598} &  61.925 &  0.012 & 0.011 & \textbf{0.008} &   0.009 &  0.969 & 0.974 & \textbf{0.983} &   0.982 \\
    5    & 56.097 & 55.129 & \textbf{56.044} &  55.703 &  0.018 & 0.019 & \textbf{0.018} &   \textbf{0.018} &  0.932 & 0.928 & \textbf{0.937} &   0.935 \\
    6    & 55.076 & 58.316 & 58.693 &  \textbf{58.758} &  0.373 & 0.255 & \textbf{0.242} &   \textbf{0.242} &  0.747 & 0.884 & 0.893 &   \textbf{0.895} \\
    7    & 40.721 & 46.084 & \textbf{46.591} &  46.496 &  1.825 & 0.972 & \textbf{0.917} &   0.932 &  0.766 & 0.899 & \textbf{0.907} &   0.905 \\
    8    & 50.669 & 57.656 & \textbf{58.432} &  58.135 &  0.613 & 0.374 & \textbf{0.226} &   0.358 &  0.747 & 0.907 & \textbf{0.913} &   0.912 \\
    9    & 47.476 & 55.287 & 56.014 &  \textbf{56.015} &  0.904 & 0.336 & 0.306 &   \textbf{0.305} &  0.756 & 0.924 & \textbf{0.929} &   \textbf{0.929} \\
    10   & 44.601 & 52.535 & \textbf{53.226} &  53.120 &  1.222 & 0.582 & \textbf{0.418} &   0.550 &  0.748 & 0.919 & \textbf{0.924} &   \textbf{0.924} \\
    11   & 38.530 & 44.753 & 45.184 &  \textbf{45.243} &  2.335 & 1.071 & \textbf{1.020} &   1.013 &  0.770 & 0.922 & \textbf{0.929} &   \textbf{0.929} \\
    12   & 43.560 & 49.568 & 50.023 &  \textbf{50.030} &  1.314 & 0.626 & \textbf{0.594} &   \textbf{0.594} &  0.762 & 0.913 & \textbf{0.921} &   0.920 \\
    13   & 38.667 & 44.925 & \textbf{45.439} &  45.343 &  2.286 & 1.177 & \textbf{0.991} &   1.130 &  0.782 & 0.925 & \textbf{0.933} &   0.932 \\
    14   & 38.167 & 44.594 & 44.996 &  \textbf{45.090} &  2.392 & 1.080 & 1.036 &   \textbf{1.024} &  0.770 & 0.924 & 0.931 &   \textbf{0.932} \\
    15   & 38.163 & 44.699 & \textbf{45.284} &  45.252 &  2.387 & 1.185 & \textbf{0.991} &   1.118 &  0.762 & 0.921 & 0.929 &   \textbf{0.930} \\
    16   & 40.578 & 47.313 & 47.731 &  \textbf{47.778} &  1.819 & 0.786 & 0.751 &   \textbf{0.745} &  0.721 & 0.892 & 0.898 &   \textbf{0.899} \\
    \bottomrule
    \end{tabular}
    \caption{Model comparison results from 200 randomly selected samples in 2019. \textbf{Bold} highlights the top performing model and * highlights the second best.}
    \label{tab:compare}

\end{table*}

\section{Experiments}

We demonstrate the effectiveness of a set of SSM models on a large-scale dataset using a high-performance computing system with a cluster of GPUs. The goal of our experiments is to show that optical flow is highly applicable for temporal interpolation of satellite imagery and compare to the baseline of linear interpolation, as traditionally applied. The following sub-sections outline the training process, compare methodologies, and study the effectiveness on a severe convective precipitation event. 

\begin{figure}
    \begin{subfigure}{0.49\linewidth}
        \centering
        \includegraphics[width=\linewidth]{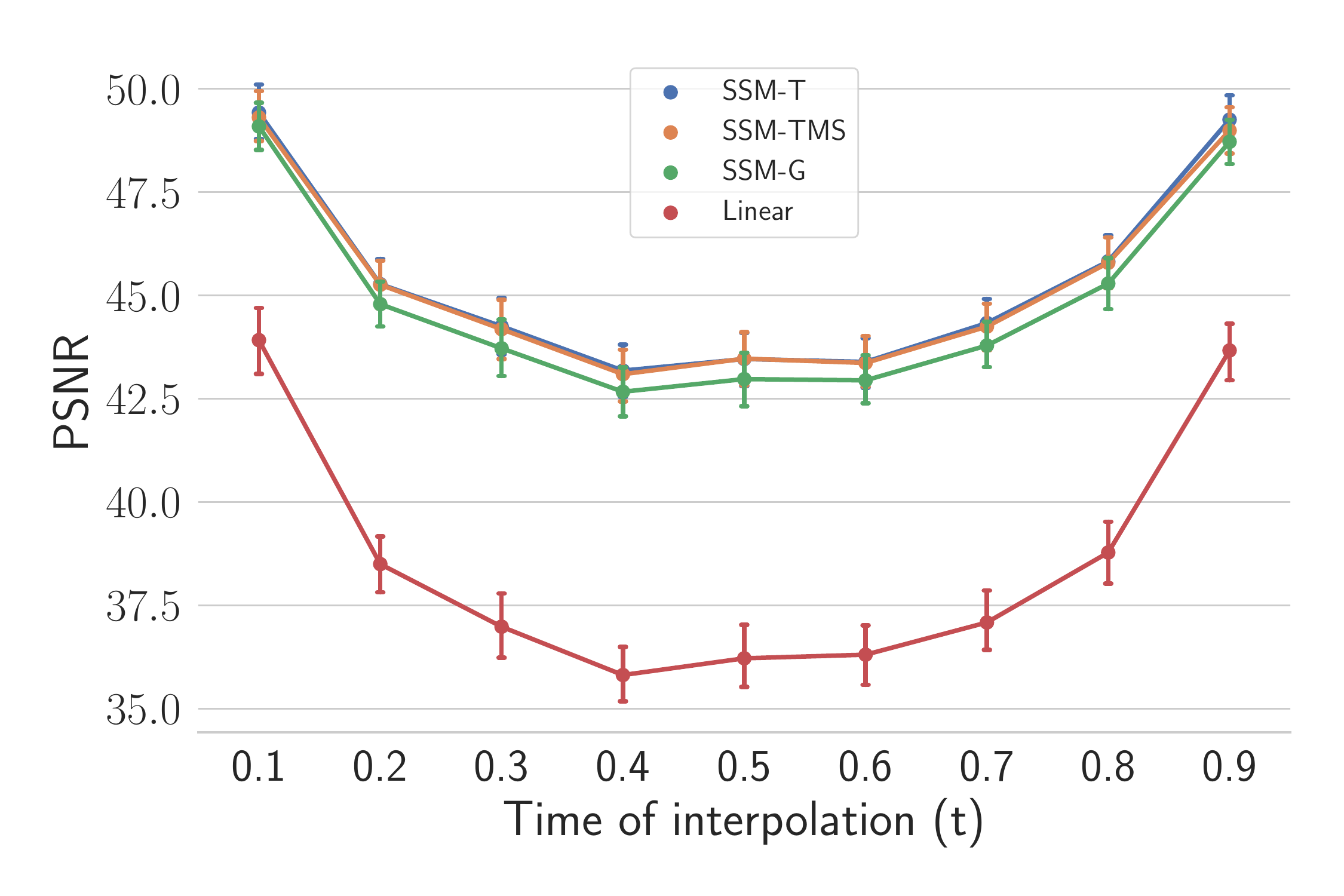}
        \caption{Interpolation Error as a function of time.}
        \label{fig:temporal_range}
    \end{subfigure}
    \begin{subfigure}{0.49\linewidth}
        \centering
        \includegraphics[width=\linewidth]{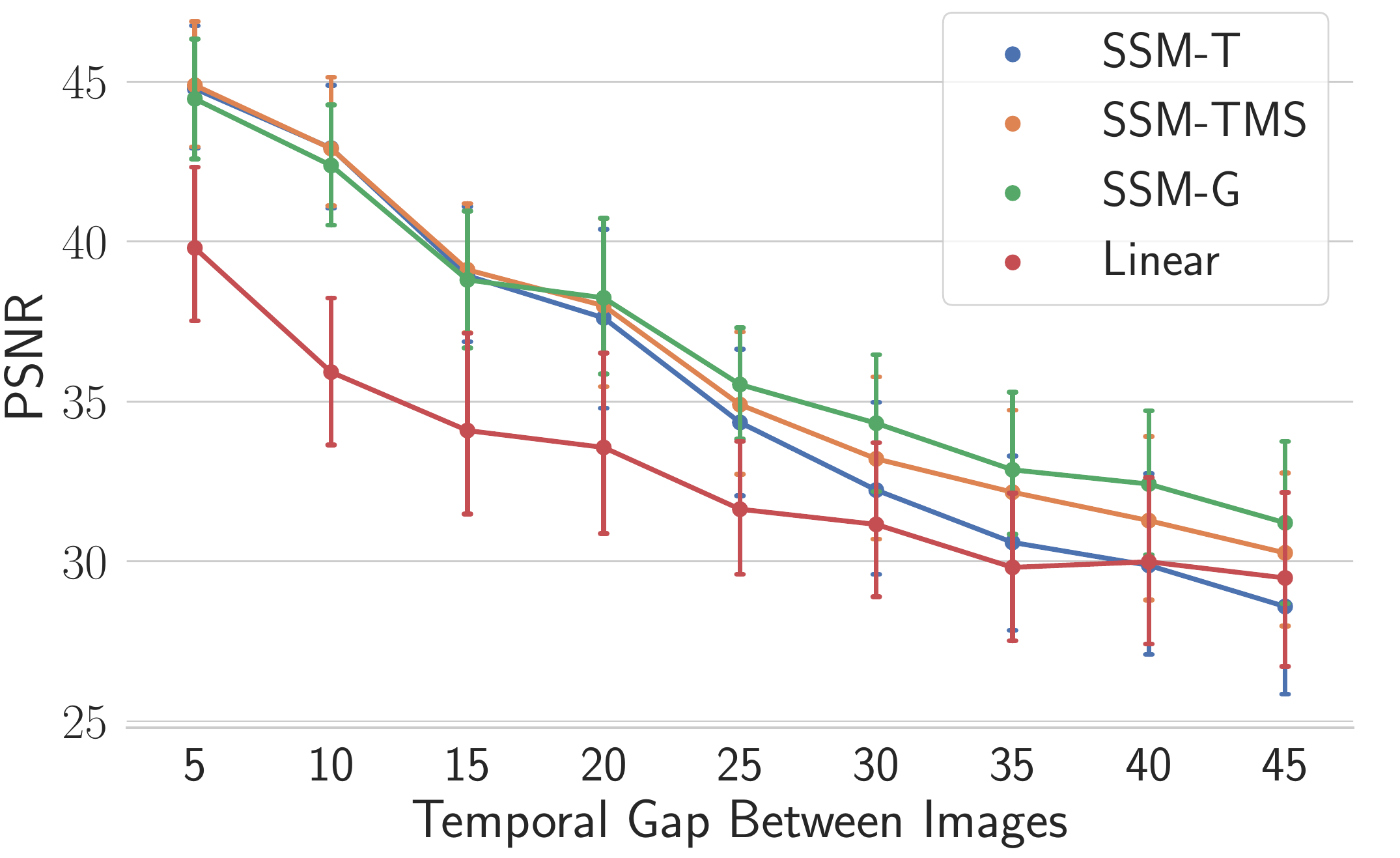}
        \caption{}
        \label{fig:temporal_gap}
    \end{subfigure}
\end{figure}

\subsection{Training}

Data for training and testing was taken from the GOES-16 Mesoscale 1-minute imagery. These images are of identical spatial and spectral resolution as North America and full-disk imagery so the learned models are directly applicable to these datasets. Training data was selected using all samples for every 5-days of the year 2018 and testing data on a randomly selected set of examples from 2019. Samples were generated as 264x264 sub-images and randomly cropped to 256x256 during training. Standardized normalization was applied independently to each channel to ensure similar pixel intensity distributions across bands. Temporally, samples are selected from a sequence of 15 time-steps such that inputs $(I_0,I_1)$ are 10-minutes apart with a random label $I_t$ in-between. Further, during training, images are randomly flipped and rotated to improve generality in the U-Net architecture.  A random training/validation split of 20\% was used to monitor learning. We select cloud top temperature tracked by band 13 ($10.3\mu m$) in ablation and demonstration experiments as used in studies of convection and atmospheric motion vectors. Experiments for this study leveraged NASA's Pleiades Supercomputer and the NASA Earth Exchange to process large-scale GOES-16 data and train individual networks for each of the 16 channels.
 
Adam optimization is used to minimize Equation~\ref{eq:totalloss} with default parameters $\beta_1=0.9$, $\beta_2=0.999$, and eps=1e-8 in PyTorch. We found that learning is sensitive to hyper-parameters $\lambda_s$ and $\lambda_w$ and are optimized using probabilistic grid search, constrained Bayesian optimization~\cite{letham2019constrained}.  Constrained Bayesian optimization applies efficient randomized Monte Carlo simulations over $\lambda_s$ and $\lambda_w$ holding $\lambda_r = 1$. We perform this process using the open-source Ax library~\cite{AxFb} for 20 trials on band 1 with SuperSlomo and find $\lambda_s=0.23$ and $\lambda_w=0.65$ minimized reconstruction loss on the validation set.  These hyper-parameters are applied to all following bands and experiments. Training the suite of models was executed on multi-node GPU cluster of V100's with 1 model per band. Multi-gpu training was used for serial hyper-parameter optimization and global models.

\input{images.tex}

\subsection{Model Comparison}

This section compares variations of SuperSlomo with a linear interpolation baseline for interpolation of geostationary images. Linear interpolation between frames is performed by taking a linear combination of two input images weighted by time, $\hat{I}_t = (1-t) * I_0 + t * I_1$. A set of three SuperSlomo models are explored including global (SSM-G), task specific (SSM-T), and task specific with multi-scale layers (SSM-TMS). We note, that due to the multi-scale blocks, SSM-TMS has fewer parameters than SSM-T. SSM-T models are trained for each band separately. SSM-G is trained using training data from all bands and hence a substantially larger training set. Root mean square error (RMSE), peak to signal noise ratio (PSNR), and self similarity measure (SSIM) are used to evaluate performance. 

We first study inherent properties of SSM on Band 13 including time dependence and sensitivity to larger displacements. Interpolation between two frames are expected to have smooth transitions from one frame to another.  Generally interpolation will have the largest error where the distance to frames is maximum (ie. directly between the input frames). In figure \ref{fig:temporal_range} we compare PSNR as a function of $t \in [0,1]$ between models and see this effect. The gap between linear and SSM models is pronounced. Between SSM models, SSM-T and SSM-TMS have similar performance. SSM-G which is a more generalized model does not perform quite as well as SSM-T and SSM-TMS, suggesting task specific models across bands perform better. Figure \ref{fig:temporal_gap} shows PSNR at $t=0.5$ while increasing the gap between $I_0$ to $I_1$ from 5 to 45-minutes. A 45-minute gap contains 9x more displacement than a 5-minute gap making the optical flow problem more difficult. Over the first 15-minutes SSM models perform similarly and better than linear. As the gap widens, SSM-TMS and SSM-G begin performing better tham SSM-T. This suggests that SSM-TMS multi-scale layers may be capturing more motion. SSM-G's more diverse dataset includes 500m data which has larger displacements than the 2-km band 13.

In table \ref{tab:compare} we present the results for each of the 16 bands of GOES-16 in 200 randomly sampled 10-minute intervals from 2019. Metrics are computed for each sample at $t=0.5$, where the error is largest, and averages over all examples. As a whole, our results find that task specific SSM models, SSM-T and SSM-TS, outperform linear interpolation and a single global interpolation network, SSM-G. Interpolation of the visible and near-infrared bands (1-6) with optical flow provided modest improvements in all metrics. Interpolation with SSM of infrared, or thermal bands, drastically improves performance on all metrics. We find that task specific models outperform the global model throughout even with the reduced training data size. SSM-T and SSM-TMS perform similarly with SSM-TMS having many fewer parameters.

\begin{figure}
    \centering
    \includegraphics[width=\linewidth]{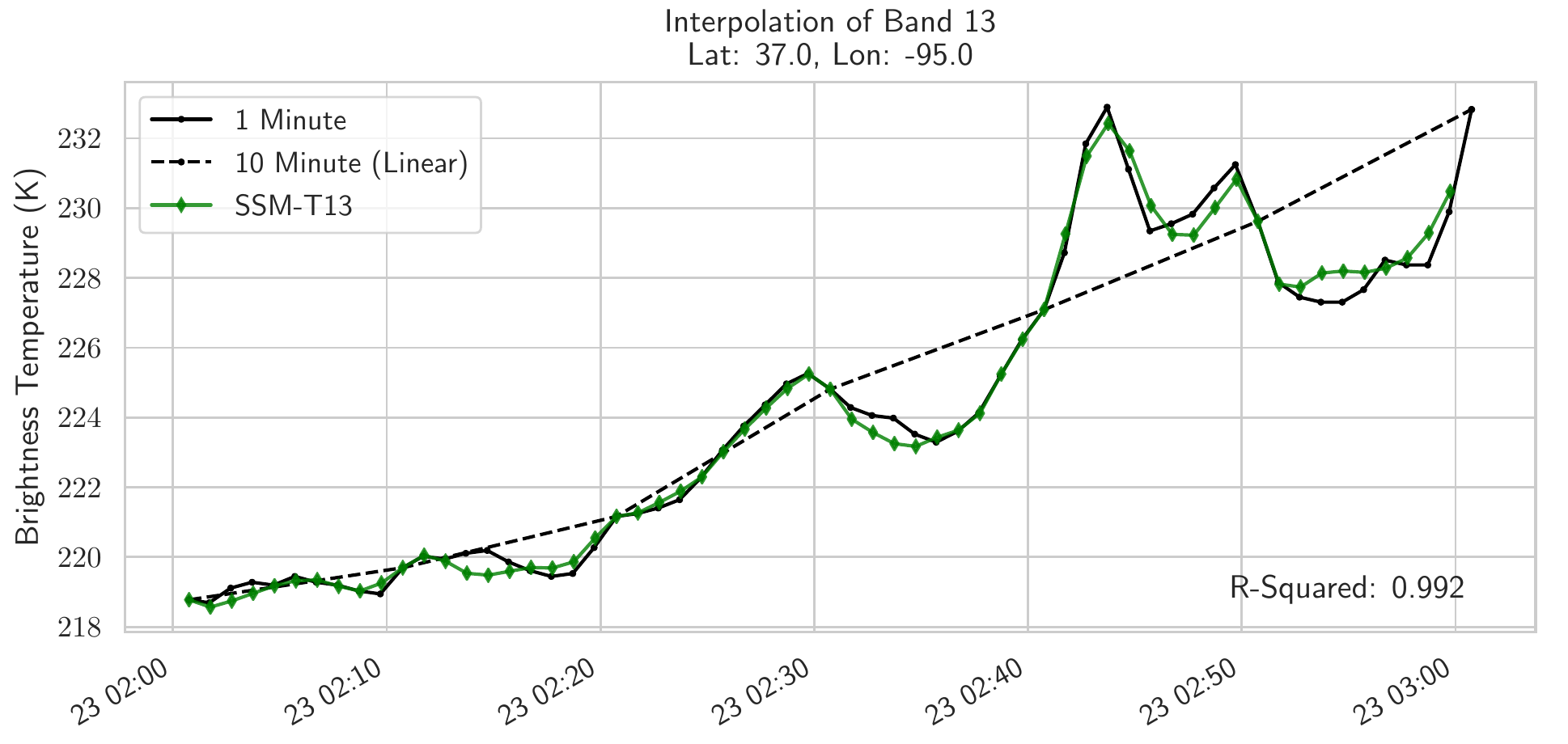}
    \caption{Cloud-top temperature during a convective event}
    \label{fig:timeseries}
\end{figure}

\subsection{Severe Weather Event}

This section studies an example of a convective precipitation event visualized in figure \ref{fig:sub:I0}. In the context of severe weather, convection is vertical motion in the atmosphere that occurs when warm air on the surfaces forces cold air in the atmosphere down often causing super-cells and heavy precipitation. For the first time, \cite{apke2016analysis} studied this process using GOES-14 1-minute imagery for a set of super-cells. The authors found that atmospheric motion can help define signatures of super-cell events to better inform weather forecasting models. Here, we show that cloud top brightness can be interpolated from 10 to 1-minute during a convective event.

One minute mesoscale (M1) data from May 23, 2019 from 2:00 to 3:00 UTC at -95$^{\circ}$ longitude and 37$^{\circ}$ latitude is used for analysis. In this region, a convective storm is occuring and moving east. The data is down-sampled to 10-minutes interpolated back to a 1-minute time-series. Figure \ref{fig:sub:I0} shows the region of interest with predictions ($I_t$), optical flows ($F_{0 \rightarrow t}$), and visibility maps ($V_{0 \rightarrow t}$) between time 0 and $t$. The optical flows show the storm moving east and slightly rotating with maximum displacement around the storm edges. According to the flows, horizontal cloud movement in the center of the storm is less than nearby areas. Visibility maps show if the corresponding pixel in $I_0$ occurs in $I_t$. Visibility pixels correspond to edges of clouds which allows \ref{eq:general-interpolation} to be a non-linear combination relative to time.

Figure \ref{fig:timeseries} presents these time-series over the defined 1-hour time-frame at (-95$^{\circ}$, 37$^{\circ}$). The time-series shows cloud top brightness increasing as warmer air rises in the atmosphere. A dashed line shows the 10-minute time-series and is equivalent to linear interpolation. SSM-T is overlayed the observation and well captures the variability of a drastic 12$^{\circ}$K temperature increase. These results suggest that optical flow may be a promising approach for interpolating geostationary imagery for applications to severe events. For reference, we include two more examples of extreme events in the supplement.

%% file: images.tex
\begin{figure*}
    \centering
        \begin{subfigure}{.19\textwidth}
          \centering
          \includegraphics[width=\linewidth]{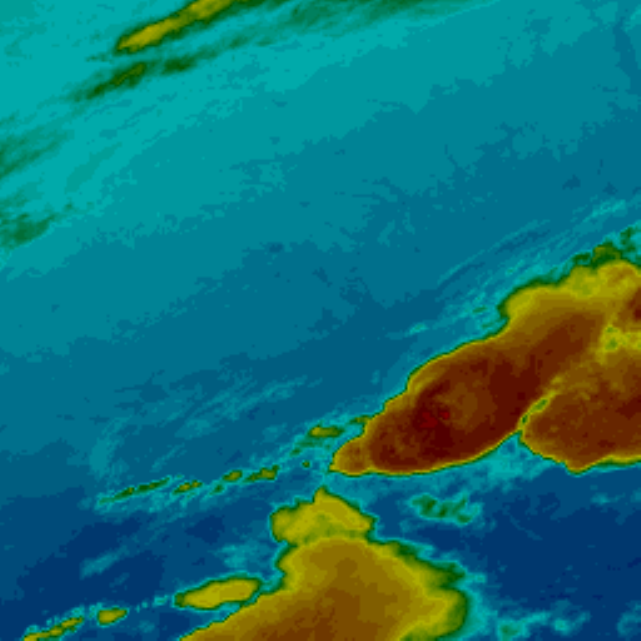}  
          \caption{$I_0$}
          \label{fig:sub:I0}
        \end{subfigure}
        \begin{subfigure}{.19\textwidth}
          \centering
          \includegraphics[width=\linewidth]{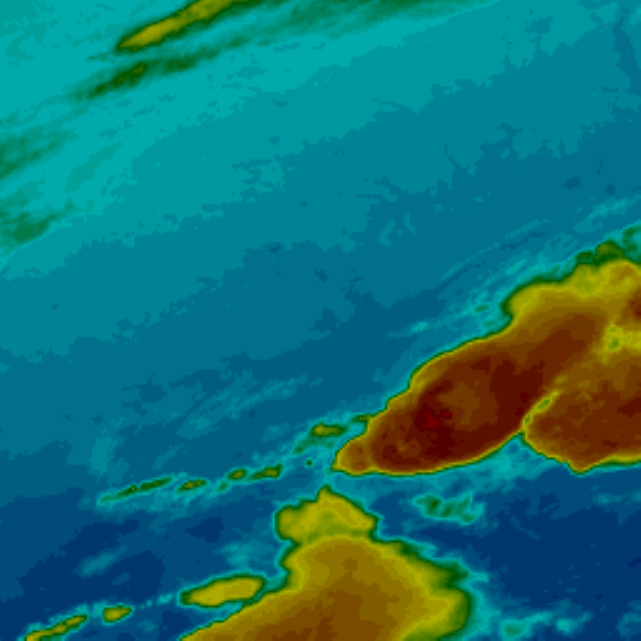}  
          \caption{$\hat{I}_{0.2}$}
          \label{fig:sub:p0.2}
        \end{subfigure}
        \begin{subfigure}{.19\textwidth}
          \centering
          \includegraphics[width=\linewidth]{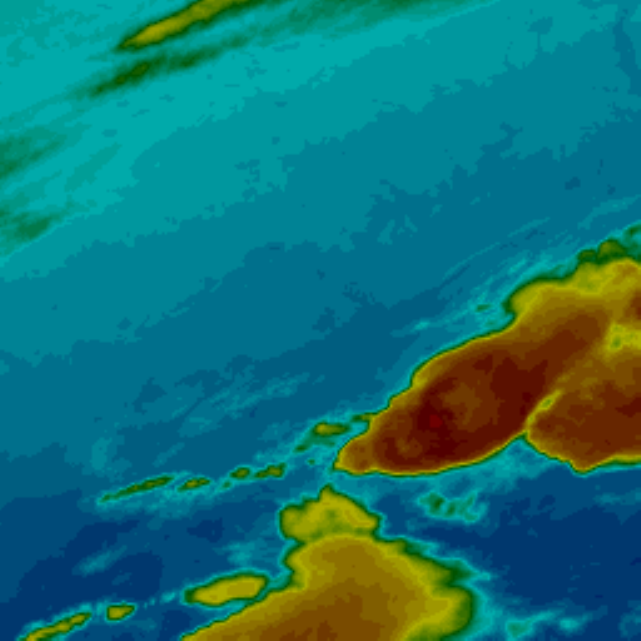} 
          \caption{$\hat{I}_{0.4}$}
          \label{fig:sub:p0.4}
        \end{subfigure}
        \begin{subfigure}{.19\textwidth}
          \centering
          \includegraphics[width=\linewidth]{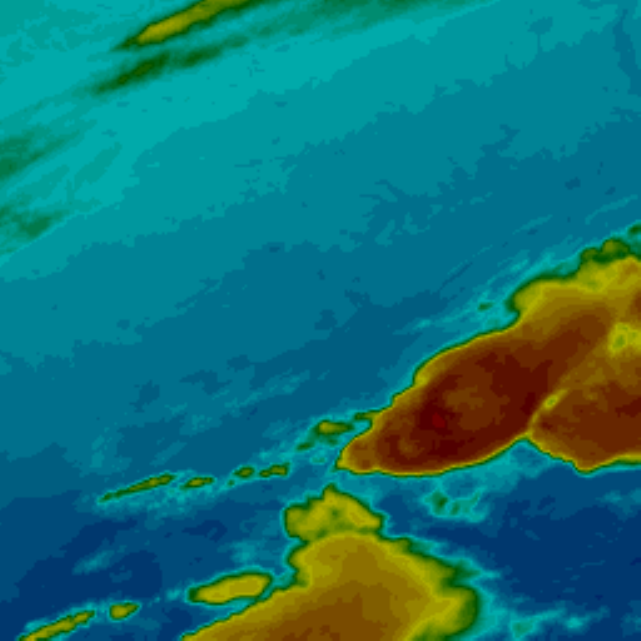}  
          \caption{$\hat{I}_{0.6}$}
          \label{fig:sub:p0.6}
        \end{subfigure}
        \begin{subfigure}{.19\textwidth}
          \centering
          \includegraphics[width=\linewidth]{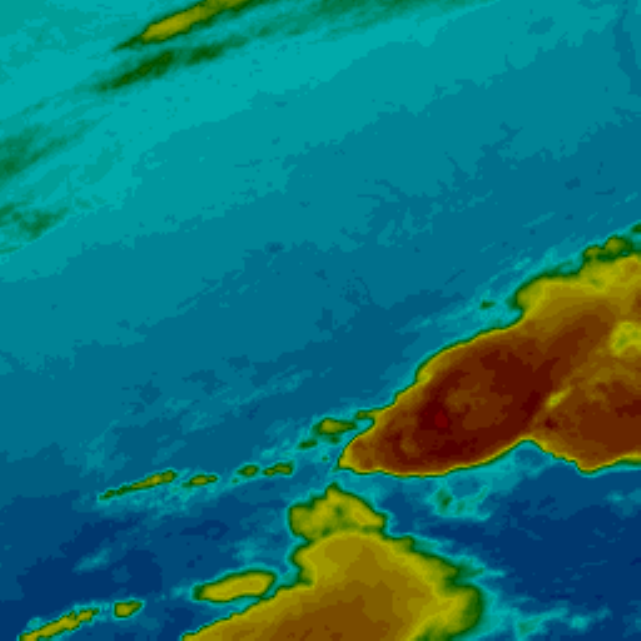}  
          \caption{$\hat{I}_{0.8}$}
          \label{fig:sub:p0.8}
        \end{subfigure}
    \newline
        \begin{subfigure}{.19\textwidth}
          \centering
          \includegraphics[width=\linewidth]{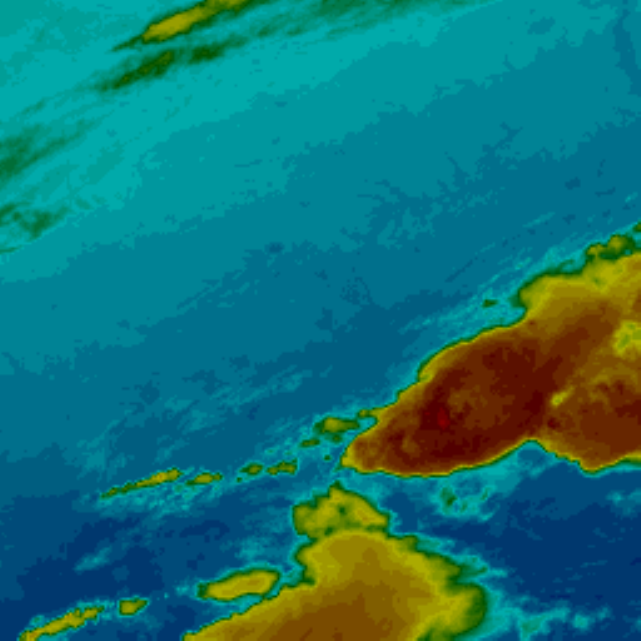}  
          \caption{$I_1$}
          \label{fig:sub:I1}
        \end{subfigure}
        \begin{subfigure}{.19\textwidth}
          \centering
          \includegraphics[width=\linewidth]{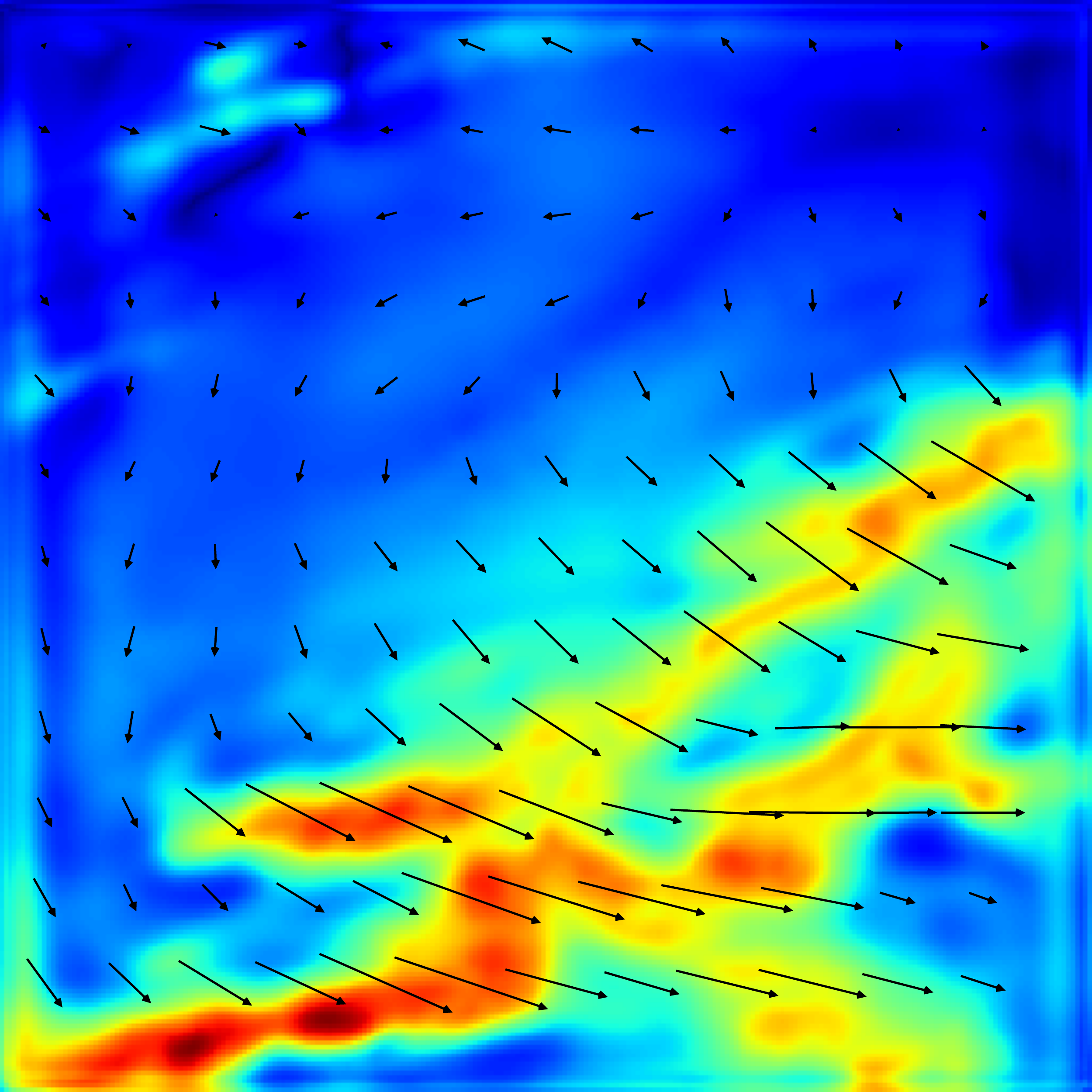}  
          \caption{$F_{0 \rightarrow 0.2}$}
          \label{fig:sub:q0.2}
        \end{subfigure}
        \begin{subfigure}{.19\textwidth}
          \centering
          \includegraphics[width=\linewidth]{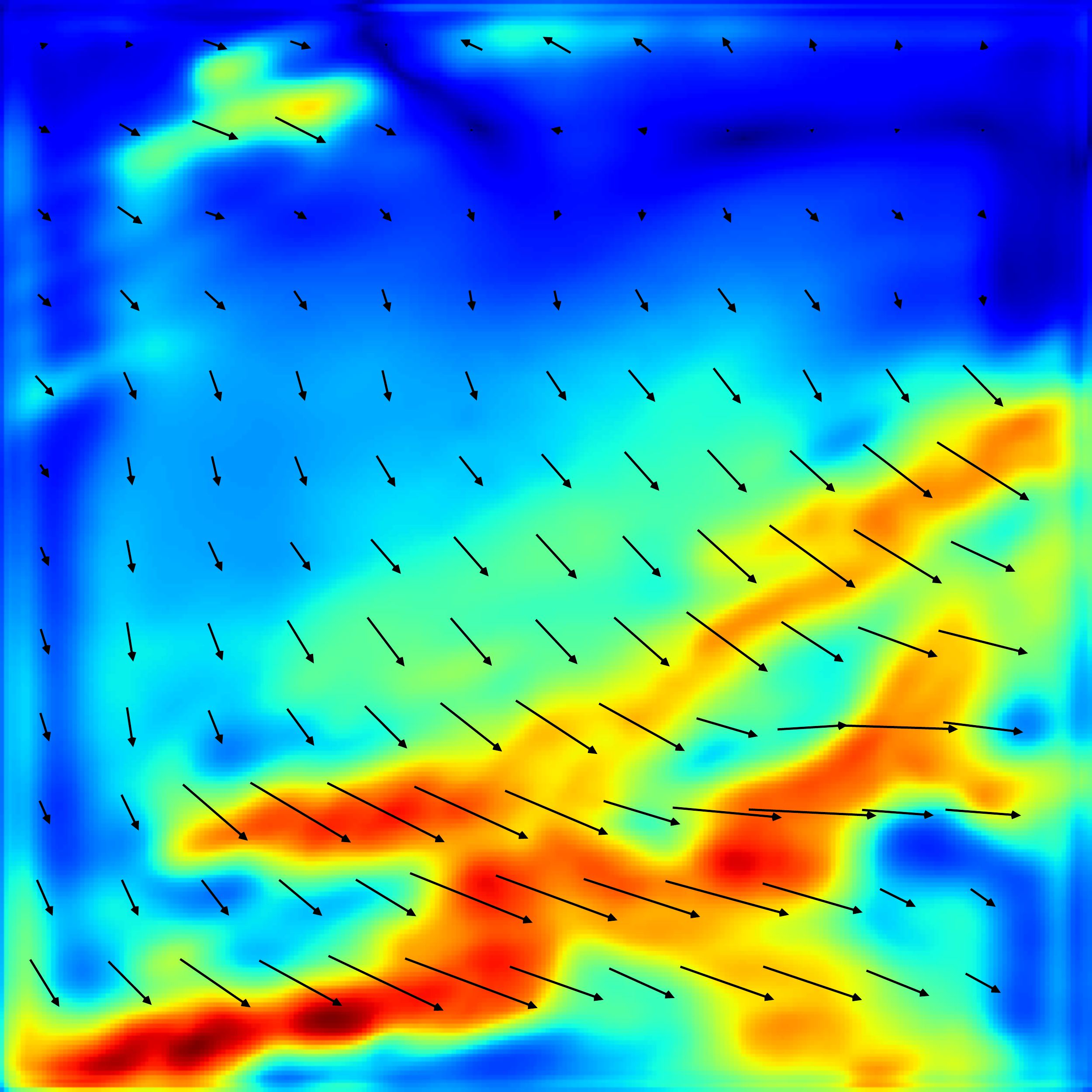} 
          \caption{$F_{0 \rightarrow 0.4}$}
          \label{fig:sub:q0.4}
        \end{subfigure}
        \begin{subfigure}{.19\textwidth}
          \centering
          \includegraphics[width=\linewidth]{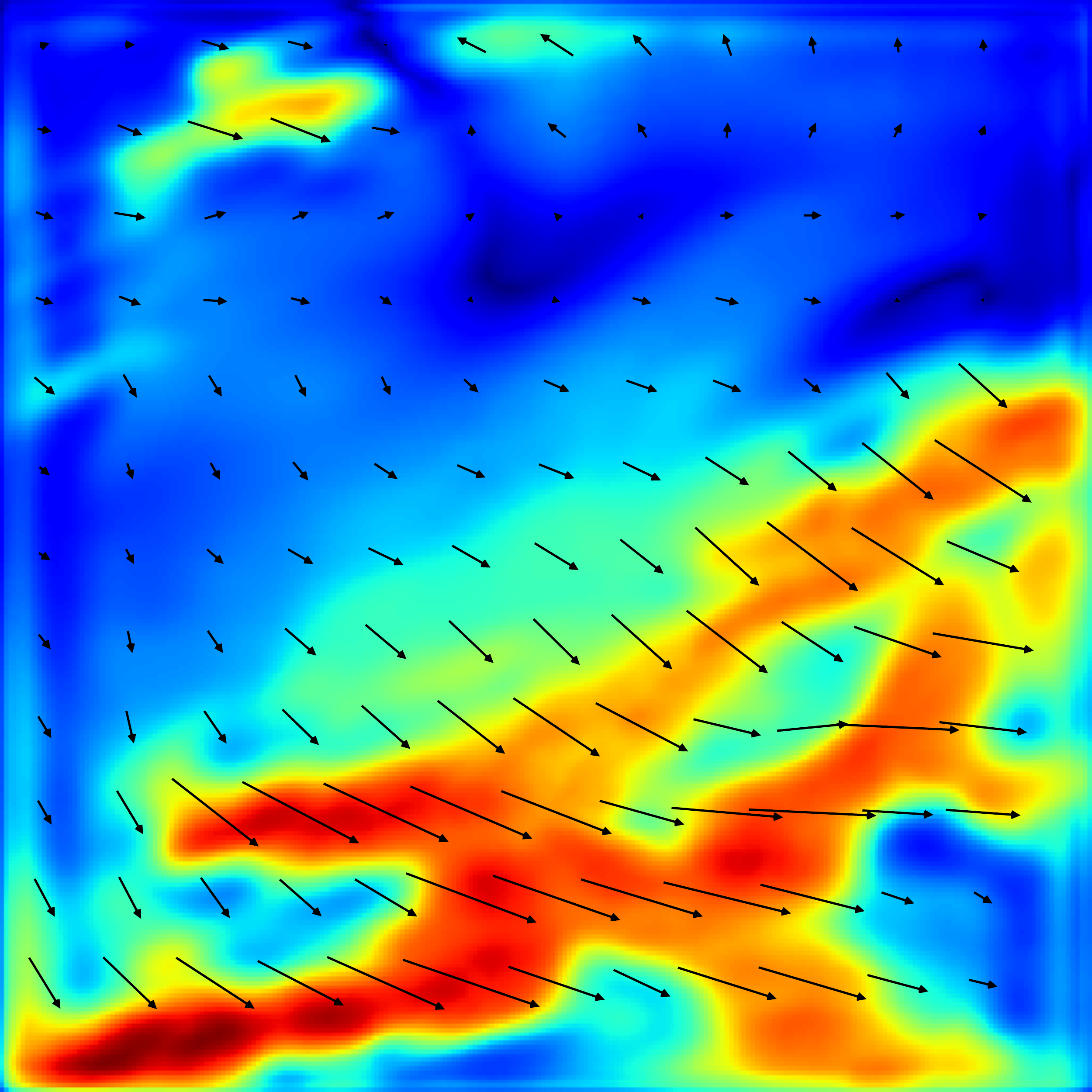}  
          \caption{$F_{0 \rightarrow 0.6}$}
          \label{fig:sub:q0.6}
        \end{subfigure}
        \begin{subfigure}{.19\textwidth}
          \centering
          \includegraphics[width=\linewidth]{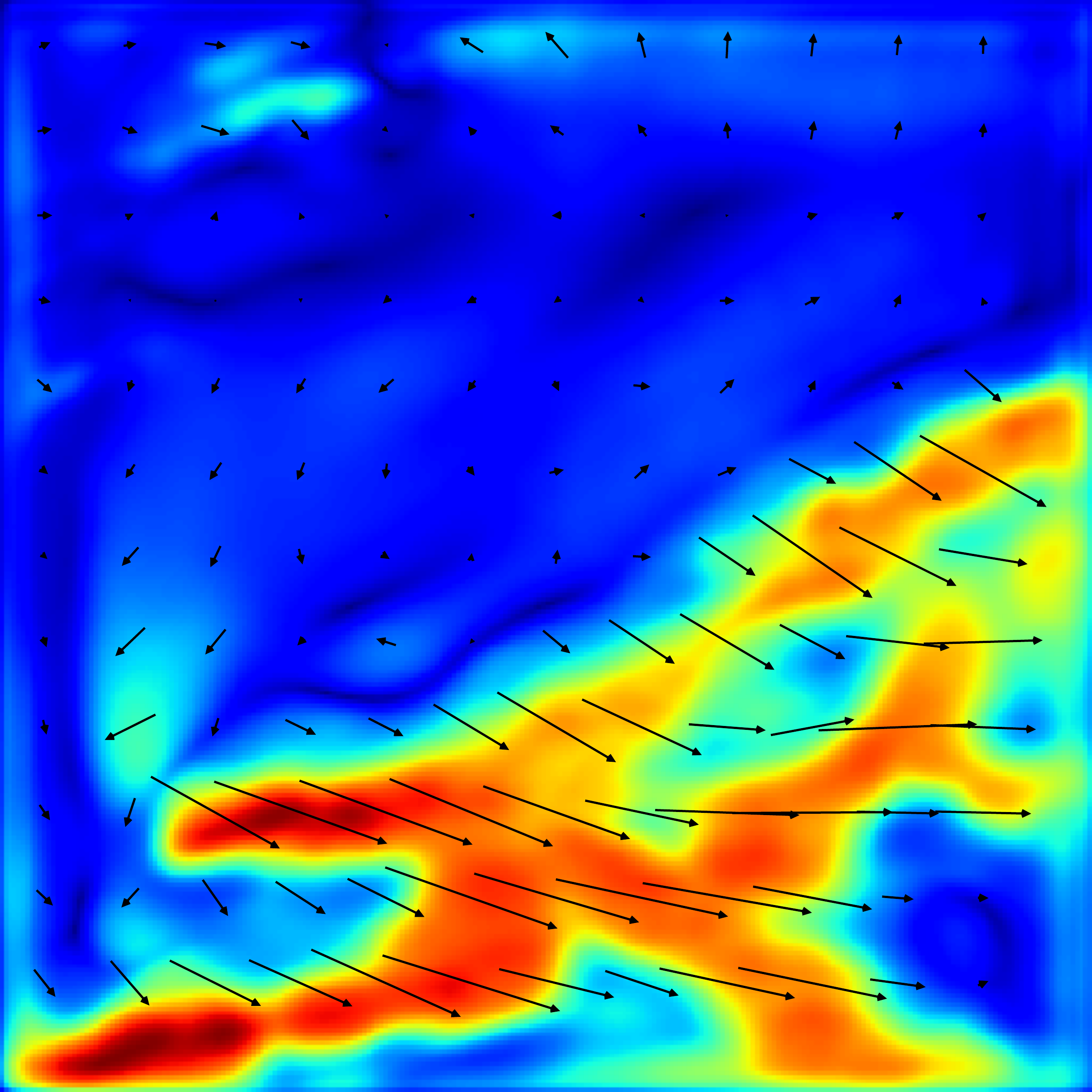}  
          \caption{$F_{0 \rightarrow 0.8}$}
          \label{fig:sub:q0.8}
        \end{subfigure}
    \newline
        \begin{subfigure}{.19\textwidth}
          \centering
          \includegraphics[width=\linewidth]{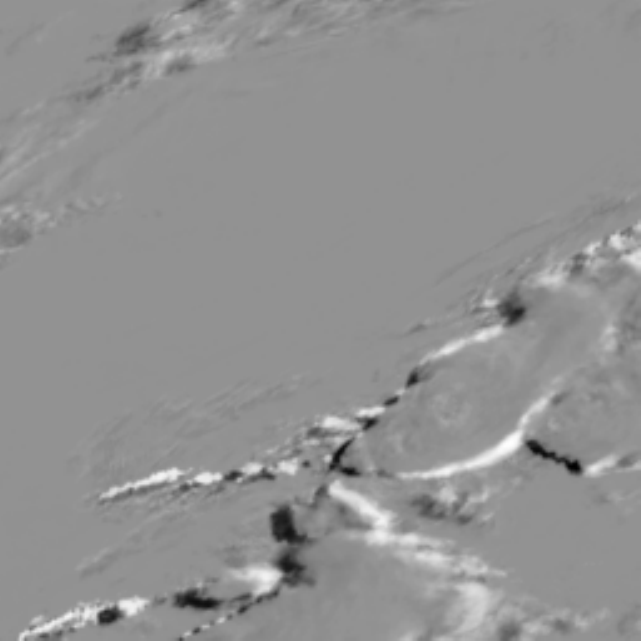}  
          \caption{$I_1-I_0$}
          \label{fig:sub:diff}
        \end{subfigure}
        \begin{subfigure}{.19\textwidth}
          \centering
          \includegraphics[width=\linewidth]{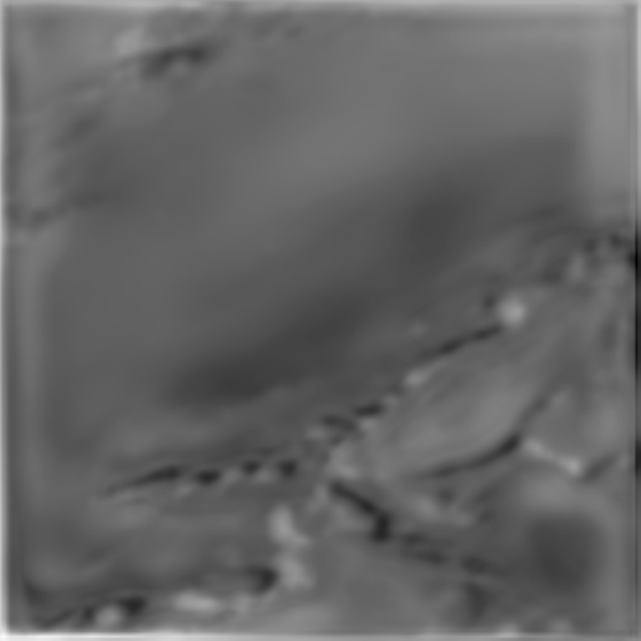}  
          \caption{$V_{0 \rightarrow 0.2}$}
          \label{fig:sub:v0.2}
        \end{subfigure}
        \begin{subfigure}{.19\textwidth}
          \centering
          \includegraphics[width=\linewidth]{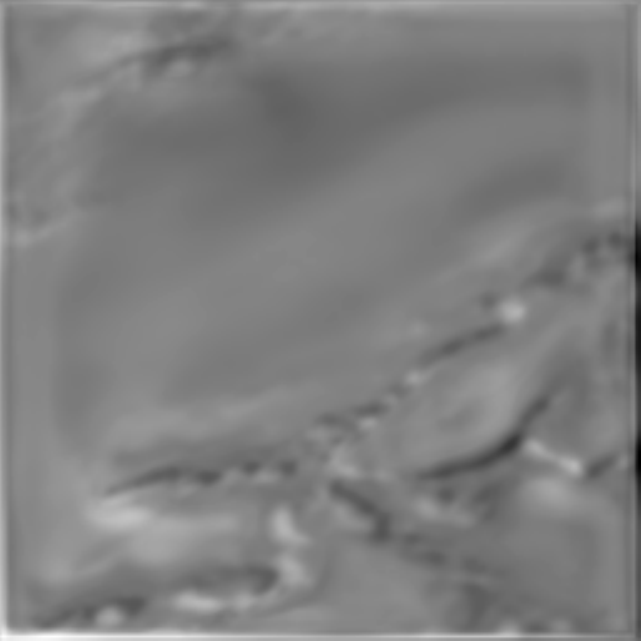} 
          \caption{$V_{0 \rightarrow 0.4}$}
          \label{fig:sub:v0.4}
        \end{subfigure}
        \begin{subfigure}{.19\textwidth}
          \centering
          \includegraphics[width=\linewidth]{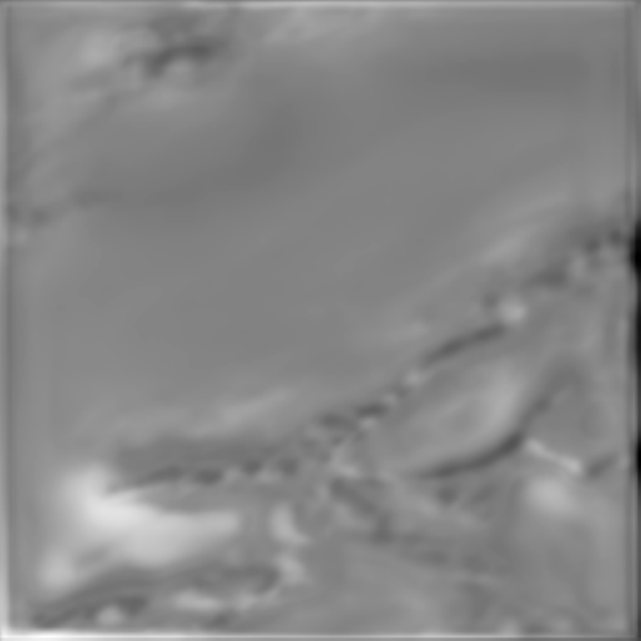}  
          \caption{$V_{0 \rightarrow 0.6}$}
          \label{fig:sub:v0.6}
        \end{subfigure}
        \begin{subfigure}{.19\textwidth}
          \centering
          \includegraphics[width=\linewidth]{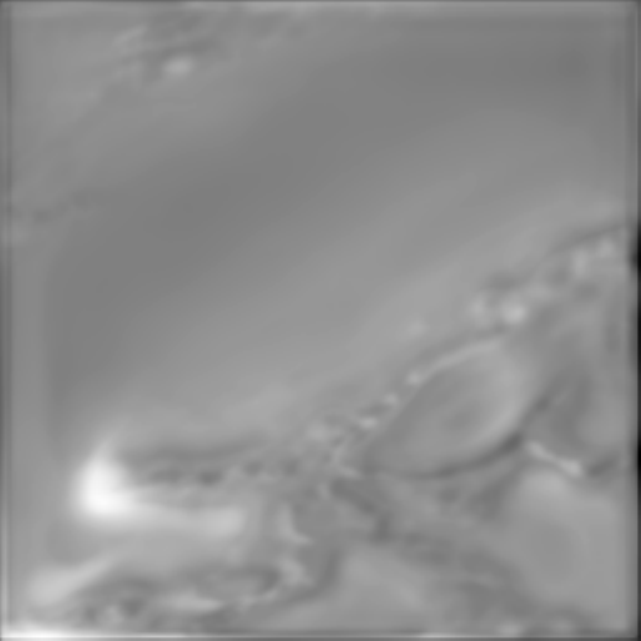}  
          \caption{$V_{0 \rightarrow 0.8}$}
          \label{fig:sub:v0.8}
        \end{subfigure}
        \caption{A severe convective event on May 23, 2019 from 2:00-2:10 UTC taken from GOES-16 Mesoscale. \ref{fig:sub:I0} and \ref{fig:sub:I1} are in the input images and their difference \ref{fig:sub:diff}. \ref{fig:sub:p0.2}-\ref{fig:sub:p0.8} show SSM-T interpolated predictions, flow intensity and direction in \ref{fig:sub:q0.2}-\ref{fig:sub:q0.8}, and visibility maps in \ref{fig:sub:v0.2}-\ref{fig:sub:v0.8}}.
\end{figure*}

%% file: conclusion.tex

\section{Conclusion}

This work proposes that temporal interpolation with optical flow is capable of modeling high-frequency events between geostationary images with high-accuracy by learning from mesoscale rapid-scan observations.  Experiments showed that learning independent weights of SSM for each band improve performance beyond one global SSM as well as the linear interpolation baseline. Multi-scale blocks in SSM-TMS have fewer parameters, performs well for larger displacements, and comparable to SSM-T overall. Interpolation well captured temporal variability of cloud top brightness during a severe convective event. This interpolation has direct applications to improved precipitation estimation and weather variability. 

While further analysis is necessary, our results suggest that dynamics of atmospheric motion is learned by the network using displacement flows and visibility maps which would have direct implications to weather forecasting. The learned optical flows derive dense atmospheric motion vectors that can be used to initialize weather models and analyze large scale winds. Secondly, internal dynamics captured may provide knowledge on how to predict future states as applied for video-frame prediction. In future work we will explore the accuracy of optical flow to estimating atmospheric motion relative to large-scale observations as well as model interpretability to better understand which physical dynamics are captured.

%% file: supplement.tex
\onecolumn
\section{Supplement}

\subsection{Hurricane Dorian - September 1, 2019}
    \begin{figure}[h!]
        \centering
        \begin{subfigure}{\linewidth}
            \centering
            \includegraphics[width=0.6\textwidth]{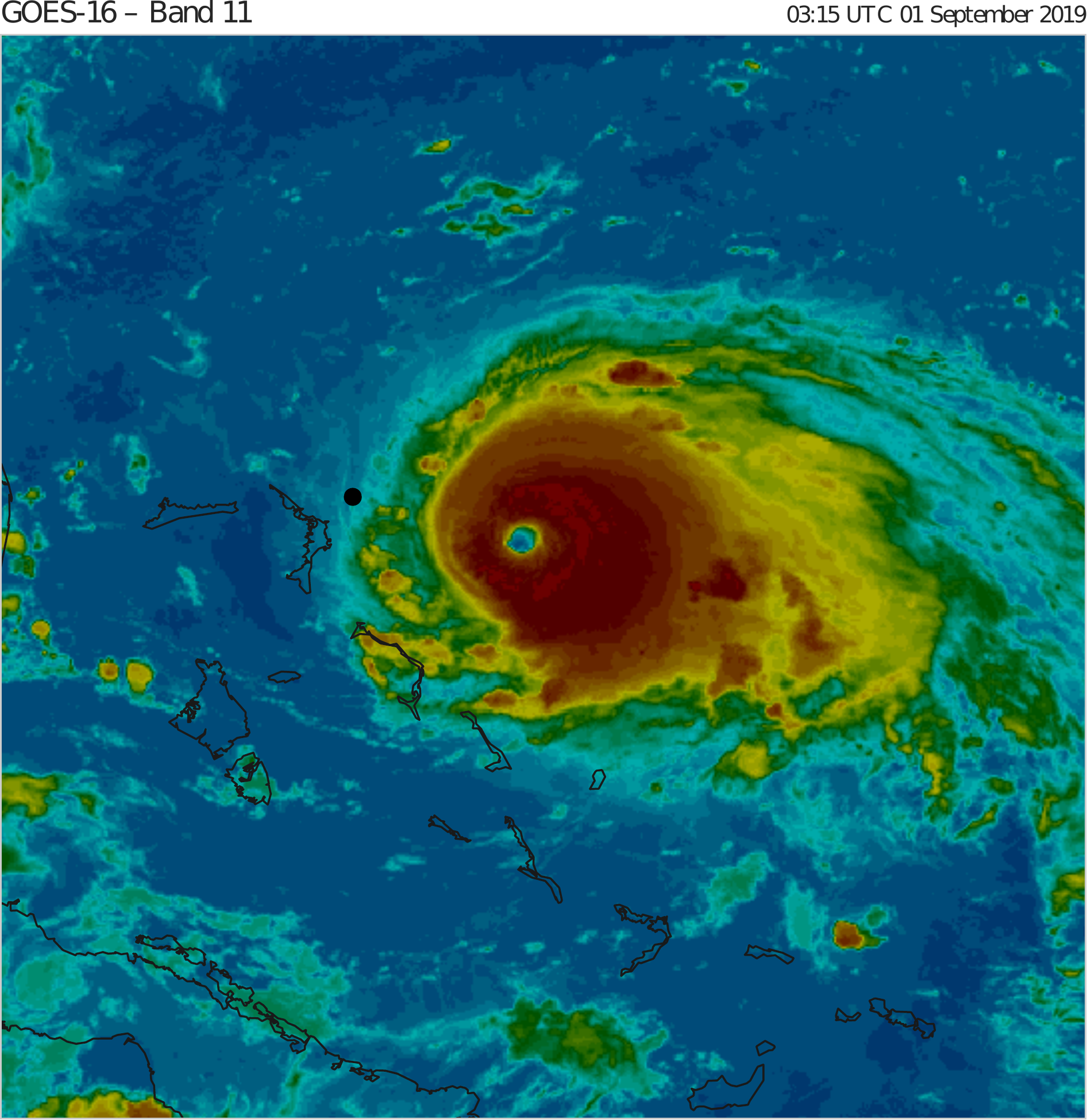}
        \end{subfigure}
        \newline
        \begin{subfigure}{\linewidth}
            \centering
            \includegraphics[width=\textwidth]{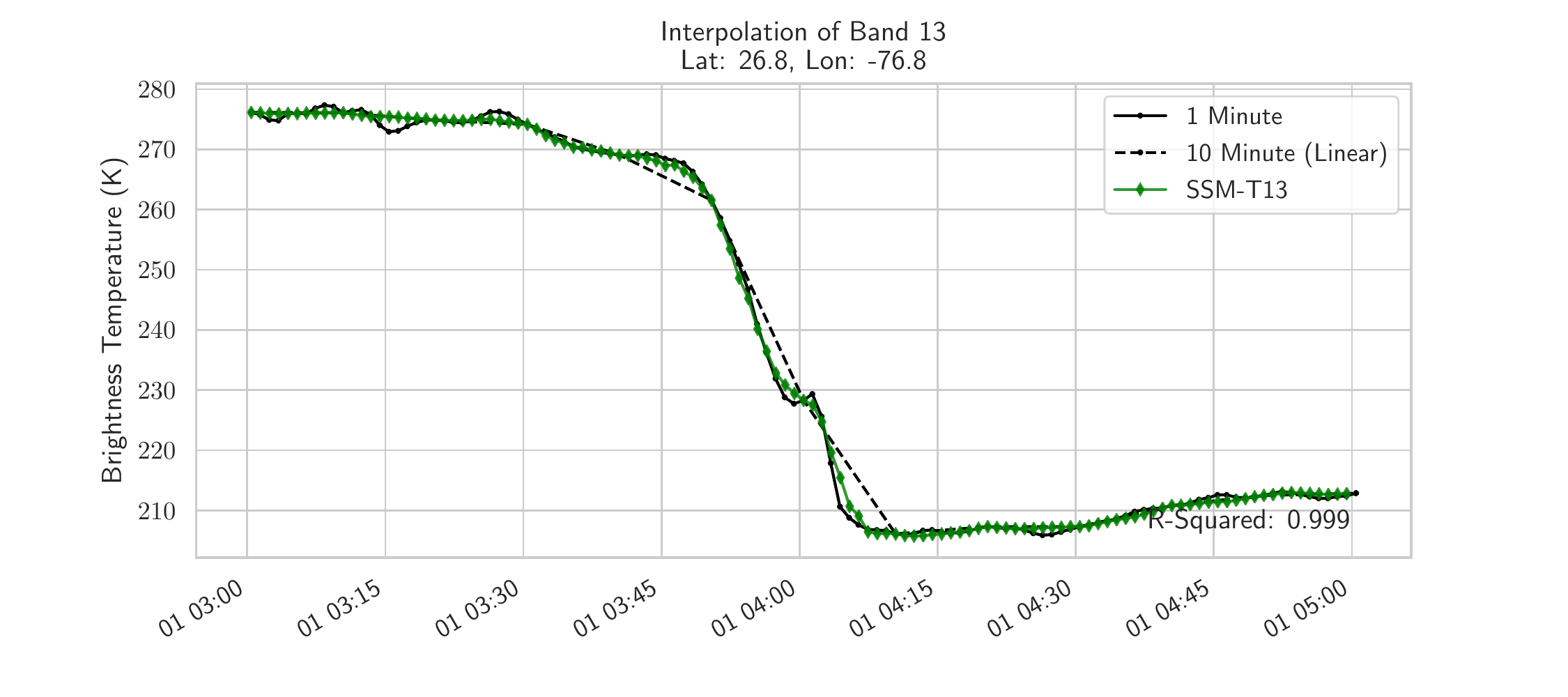}
        \end{subfigure}
        \caption{Hurricane Dorian (Category 5) on September 1, 2019.}
        \label{fig:supp1}
    \end{figure}

\newpage 
\subsection{Regional Tornado Outbreak - March 3, 2019}
    \begin{figure}[h!]
        \centering
        \begin{subfigure}{\linewidth}
            \centering
            \includegraphics[width=0.6\textwidth]{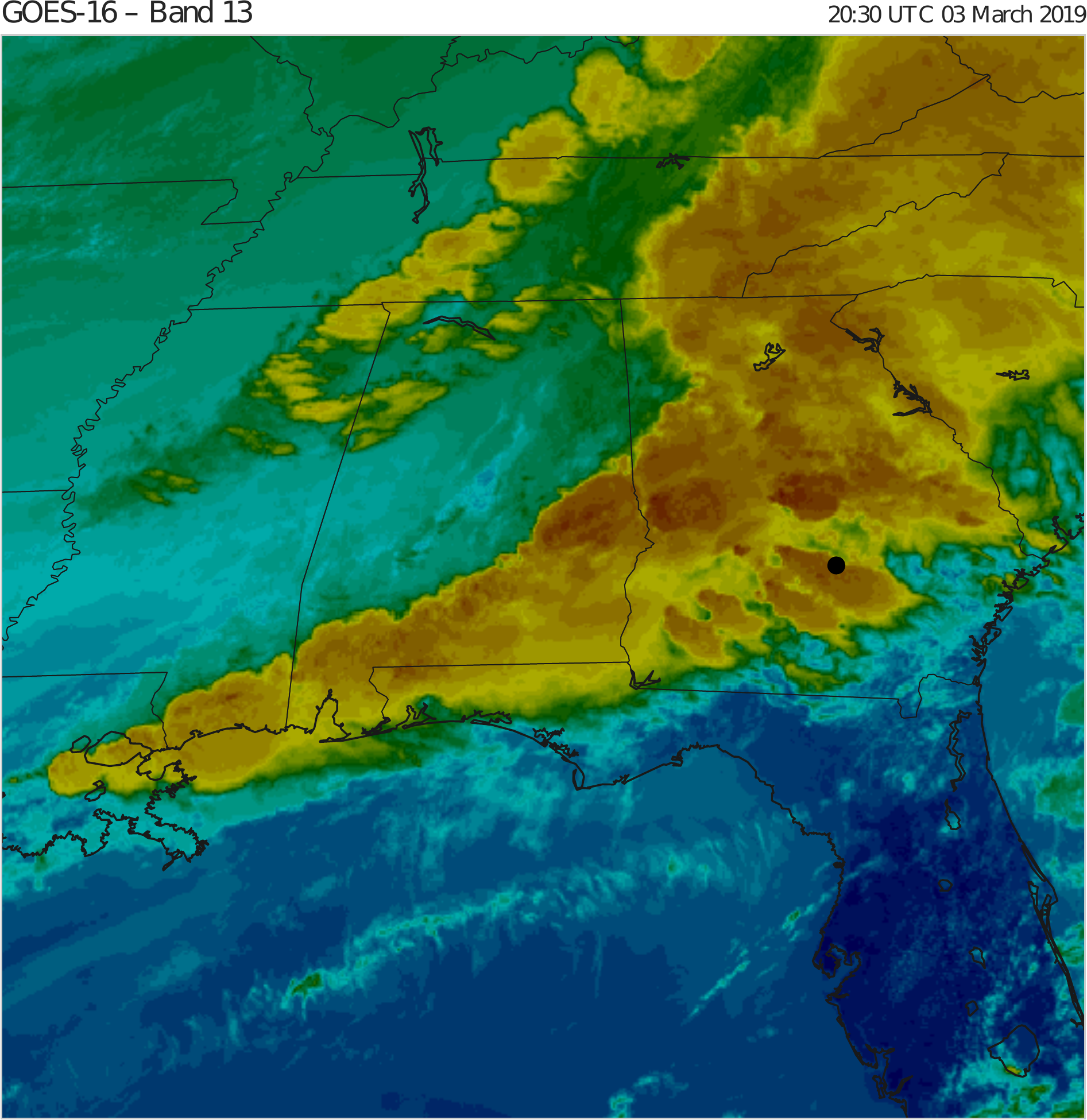}
        \end{subfigure}
        \newline
        \begin{subfigure}{\linewidth}
            \centering
            \includegraphics[width=\textwidth]{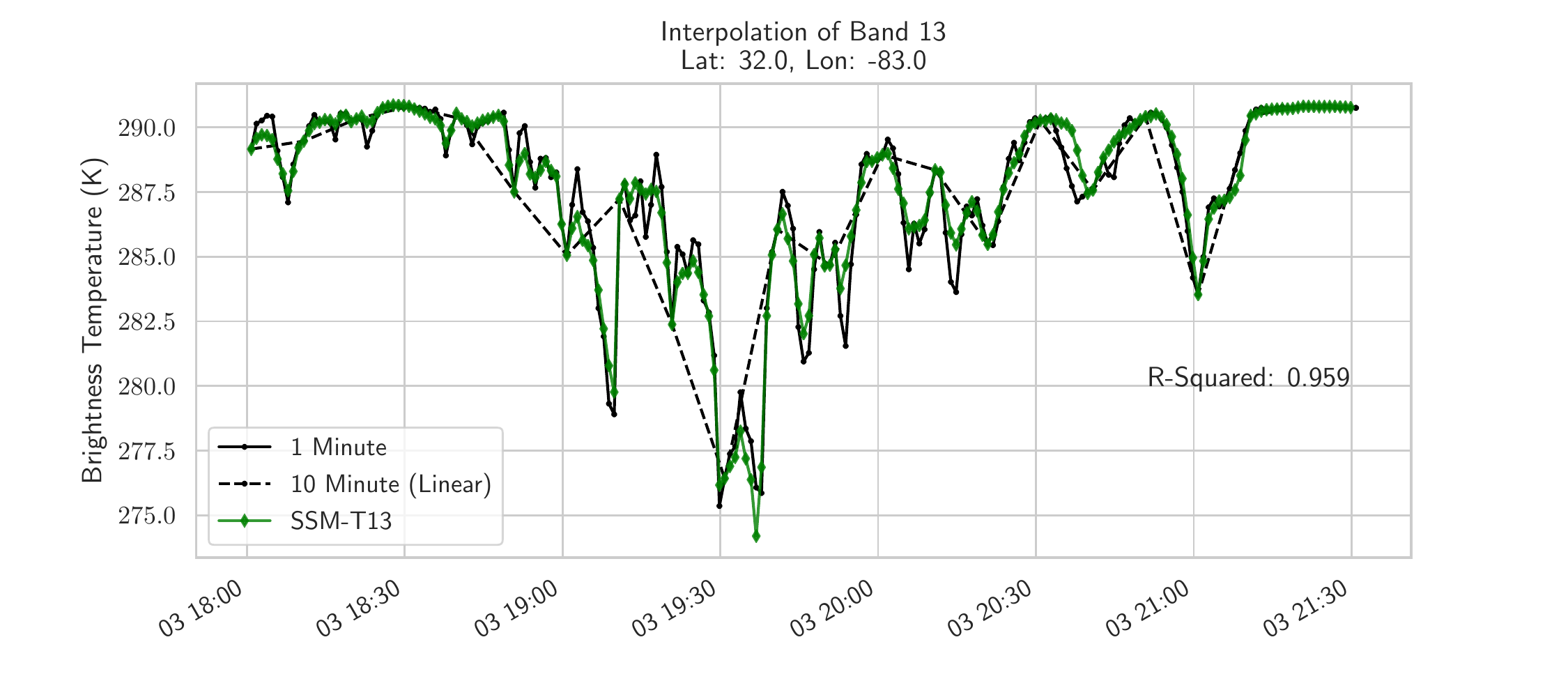}
        \end{subfigure}
        \caption{Tornado outbreak on March 3, 2019 in the Southeastern United States.}
        \label{fig:supp1}
    \end{figure}